%% file: main.tex
\documentclass[]{oppo_arxiv}

\usepackage[toc,page,header]{appendix}
\usepackage{fancyvrb}
\usepackage{wrapfig}
\usepackage[authoryear]{natbib}
\usepackage[percent]{overpic}

\input{common}

\title{A\textsuperscript{2}FM: An Adaptive Agent Foundation Model for Tool-Aware Hybrid Reasoning}

\affiliation{OPPO AI Agent Team}

\input{sec1_abstract}

\date{\today}
\checkdata[Open Source]{%
  \href{https://github.com/OPPO-PersonalAI/Adaptive_Agent_Foundation_Models}{%
    \faGithub \ Code%
  }
  \
  \href{https://huggingface.co/PersonalAILab/A2FM-32B-rl}{%
    \raisebox{-.15em}{\includegraphics[height=1em]{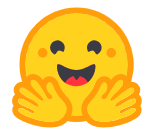}} Models%
  }
  \
}
\correspondence{Wangchunshu Zhou at \email{zhouwangchunshu@oppo.com}}

\begin{document}
\maketitle

\begin{figure}[htbp]
  \centering
  \includegraphics[width=\textwidth]{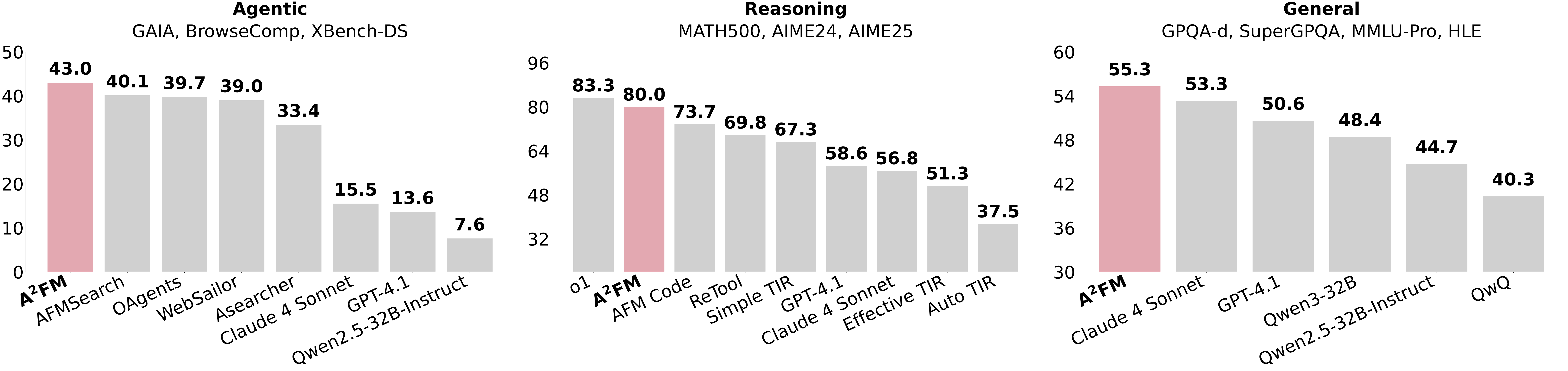}
  \caption{Average performance on agentic, reasoning, and general (ARG) benchmarks. Overall, A$^2$FM ranks 1st, 2nd, and 1st on the three categories, respectively. Note that we uses the best-suited mode for each dataset respectively.}
  \label{fig:compare_main}
\end{figure}

\clearpage
\setcounter{tocdepth}{3}

\renewcommand{\contentsname}{\LARGE\bfseries Contents}

\renewcommand{\cftsecfont}{\large\bfseries}
\renewcommand{\cftsecpagefont}{\large\bfseries}

\renewcommand{\cftsubsecfont}{\normalfont\itshape\normalsize}
\renewcommand{\cftsubsecpagefont}{\normalfont\itshape\normalsize}

\renewcommand{\cftsubsubsecfont}{\normalfont\itshape\normalsize}
\renewcommand{\cftsubsubsecpagefont}{\normalfont\itshape\normalsize}

\setlength{\cftbeforesecskip}{5pt}
\setlength{\cftbeforesubsecskip}{3pt}
\setlength{\cftbeforesubsubsecskip}{2pt}

\tableofcontents
\clearpage

\input{sec2_intro}

\input{sec3_related}

\input{sec4_method}
\input{sec5_exp}
\input{sec6_analysis}

\input{sec7_conclusion}

\input{ack}
\clearpage
\bibliography{main}
\bibliographystyle{apalike}

\newpage
\appendix

\input{sec9_app}

\end{document}

%% file: common.tex
\usepackage{times}
\usepackage{latexsym}
\usepackage[T1]{fontenc}
\usepackage[utf8]{inputenc}
\usepackage{microtype}
\usepackage{inconsolata}
\usepackage{graphicx}
\usepackage{hyperref}       
\usepackage{url}            
\usepackage{booktabs}       
\usepackage{amsfonts}       
\usepackage{nicefrac}       
\usepackage{stackengine}
\usepackage{microtype}      
\usepackage{colortbl}
\usepackage{xcolor}
\usepackage{amsmath}
\usepackage{amssymb}
\usepackage{amsthm}
\usepackage{mathrsfs}
\usepackage{pifont}
\usepackage{MnSymbol}
\usepackage{balance}
\usepackage{enumitem}
\usepackage{xcolor}
\usepackage{natbib}
\usepackage{multicol}
\usepackage{xspace}
\usepackage{fancyvrb}
\usepackage{geometry}
\usepackage{tocloft}
\usepackage{tasks}
\usepackage{hyperref}
\usepackage{fontawesome5}
\usepackage{makecell}
\AtBeginDocument{%
  \providecommand\BibTeX{{%
    \normalfont B\kern-0.5em{\scshape i\kern-0.25em b}\kern-0.8em\TeX}}}

\makeatletter
\DeclareRobustCommand\onedot{\futurelet\@let@token\@onedot}
\def\@onedot{\ifx\@let@token.\else.\null\fi}

\usepackage{setspace}
\usepackage{mathtools}

\usepackage{multirow,booktabs}
\usepackage{subcaption}

%% file: sec1_abstract.tex
\abstract{
Large language models split into two families: reasoning-centric LLMs, which strengthen internal chain-of-thought reasoning but cannot invoke external tools, and agentic LLMs, which learn to interact with environments and leverage tools but often lag in deep reasoning. This divide arises from fundamentally different training objectives, leading to mismatched strengths and inefficiency on simple queries, where both families tend to overthink or over-call tools. In this work, we present Adaptive Agent Foundation Model (A\textsuperscript{2}FM), a unified framework that follows a route-then-align principle: the model first learns task-aware routing and then aligns mode-specific trajectories under a shared backbone. To address the inefficiency gap, we introduce a third mode-instant-that handles simple queries directly, preventing unnecessary reasoning or tool calls while complementing the agentic and reasoning modes. To jointly enhance accuracy and efficiency, we propose Adaptive Policy Optimization (APO), which enforces adaptive sampling across modes and applies a cost-regularized reward. On the 32B scale, A\textsuperscript{2}FM achieves 13.4\% on BrowseComp, 70.4\% on AIME25, and 16.7\% on HLE, setting new SOTA among comparable models and performing competitively with frontier LLMs across agentic, reasoning, and general benchmarks. Notably, the adaptive execution achieves a cost of pass of only \$0.00487 per correct answer-cutting cost by 45.2\% relative to reasoning and 33.5\% relative to agentic, thus delivering substantially higher cost efficiency while maintaining comparable accuracy.
}

%% file: sec2_intro.tex
\section{Introduction}

Recent reasoning-centric Large Language Models (LLMs)—e.g., OpenAI o3~\citep{openai2025o3} and DeepSeek-R1~\citep{guo2025deepseekr1}—show strong chain-of-thought (CoT) abilities for mathematics, scientific diagram interpretation, and logic. Their post-training typically optimizes internal deliberation and solution steps. By design, they are trained in closed, text-only settings and do not learn to invoke external tools. This difference in training objective and environment leaves a capability gap in tool use.

\begin{figure}[ht]
    \centering
    \includegraphics[width=\linewidth]{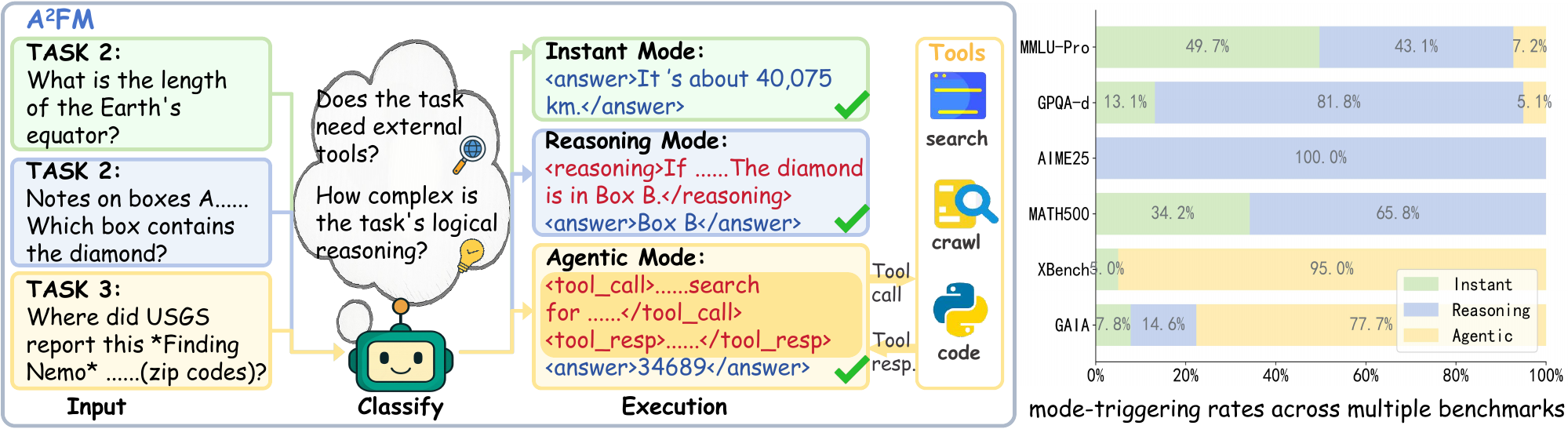}
    \caption{Overview of A\textsuperscript{2}FM.  Left: the framework integrates three execution modes—\textit{instant}, \textit{reasoning}, and \textit{agentic}—under a unified backbone with task-aware routing. Right: mode allocation across six benchmarks (MMLU-Pro, GPQA-d, AIME25, MATH500, Xbench-DeepSearch (Xbench), and GAIA-text (GAIA), illustrating how A\textsuperscript{2}FM adapts routing to task characteristics.}
    \label{fig:adaptive_afm_1}
    \vspace{-8pt}
\end{figure}
In parallel, agentic LLMs—e.g., OpenAI GPT-5\citep{openai2025gpt5}, GLM-4.5~\citep{zeng2025glm}, Kimi K2~\citep{team2025kimik2}, DeepSeek-V3.1~\citep{deepseekai2025deepseekv31}, and agent frameworks such as OAgents~\citep{zhu2025oagents} and BrowseMaster~\citep{pang2025browsemaster}, as well as Agent Foundation Models~\citep{li2025coa}—emphasize environment-dependent interaction. They plan, search, crawl, and execute code. Their post-training focuses on tool correctness, call patterns, and interaction quality. Yet when deep multi-step reasoning is required (e.g., competitive math), these systems often trail state-of-the-art reasoning models~\citep{jimenez2023swe}. Thus, the two families exhibit complementary strengths: internal CoT versus external tool use.

Recent advances in agent systems and frameworks~\citep{zhou2023agents,zhou2024agents2,zhu2025oagents,li2025websailor,wu2025webdancer,tao2025webshaper,smolagents,tang2025agent} have greatly expanded the scope of LLM-based autonomy, enabling complex behaviors such as planning, multi-step tool use, and collaborative decision-making. These systems often rely on external orchestration—manually designed workflows, multi-agent coordination, or prompt-based tool pipelines—to decompose tasks and integrate environment feedback. Despite their impressive flexibility, such designs face several inherent limitations.  
First, they incur significant computational and latency overhead due to redundant inter-agent communication and long reasoning traces. Second, their generalization ability is limited: adapting to new domains typically requires human-engineered prompt flows or redefined agent roles~\citep{zeng2023flowmind,zhang2024aflow}. Third, these frameworks depend on general-purpose LLMs that are not trained end-to-end to handle multi-turn, multi-tool reasoning, preventing them from learning data-driven strategies for mode switching.

Complementary to the above, recent works attempt to blend “thinking’’ and “non-thinking’’ behaviors to reduce redundant computation~\citep{chen2024not,fatemi2025concise,lou2025adacot,kumar2025overthink,sui2025stop}. However, these approaches generally operate along a single binary axis—deciding whether to reason or not—without the capacity to adaptively invoke \textit{agentic} behaviors such as tool use and task planning. As a result, while they can shorten over-extended reasoning traces, they remain unable to manage the broader spectrum of reasoning–acting trade-offs inherent to real-world tasks. Moreover, introducing agentic behaviors further compounds this limitation by increasing runtime and inflating the overall \textit{cost of pass}~\citep{erol2025cost}. 

To address the challenges mentioned above, we propose \textbf{A\textsuperscript{2}FM (Adaptive Agent Foundation Model)}, which unifies three complementary modes within a single backbone - \textit{agentic} (tool-aware action), \textit{reasoning} (explicit CoT), and \textit{instant} (direct answer) -  thereby bridging the gap between reasoning-centric
LLMs and orchestrated agent systems. To bridge the capability gap, A\textsuperscript{2}FM internalizes a self-adaptive router that learns \textit{what to do} per query, naturally combining the strengths of reasoning and agentic paradigms in one model. To address efficiency, we add the \textit{instant} mode so that simple queries are answered directly, avoiding unnecessary reasoning or tool calls. Supervised fine-tuning follows a \textit{route-then-align} principle: the model first performs task-aware routing, then aligns mode-conditioned trajectories under a shared backbone using diverse datasets spanning tool use, explicit reasoning, and direct answering.

Nevertheless, simply mixing modes falls short: the model must not only maintain high accuracy but also minimize computational cost, yet boundary queries remain difficult to route and data often underutilized. Prior work either omits explicit modeling of mode mixing~\citep{jiang2025thinkneedlargehybridreasoning, zhang2025adaptthink}, or focuses solely on efficiency by shortening reasoning traces in post-training~\citep{team2025kimi, arora2025training}, or fails to jointly optimize routing and trajectory generation in an end-to-end framework~\citep{wu2025unlocking}. We therefore introduce \textbf{Adaptive Policy Optimization (APO)}, a reinforcement learning procedure tailored for mode selection. APO builds on group-relative policy optimization~\citep{shao2024deepseekmath_grpo}, enforces adaptive sampling across modes to prevent collapse, and applies a dynamic, cost-regularized reward (with LLM-as-Judge signals~\citep{zheng2023llmjudge}) that balances accuracy with efficiency. It explicitly favors minimal-effort \textit{instant} solutions on easy queries while escalating to \textit{reasoning} or \textit{agentic} modes when external evidence or extended deliberation is needed.

Empirically, A\textsuperscript{2}FM achieves state-of-the-art results across benchmarks: on agentic tasks it obtains 13.4\% on BrowseComp, on reasoning it reaches 70.4\% on AIME25, and on general tasks it delivers 16.7\% on HLE. Beyond raw accuracy, APO substantially improves efficiency: on SuperGPQA, the adaptive execution yields a cost of pass of only \textbf{\$0.00487} per correct answer—cutting cost by \textbf{45.2\%} relative to reasoning and \textbf{33.5\%} relative to agentic—while maintaining comparable accuracy.

In summary, our key contributions include:
\begin{itemize}[leftmargin=*]
    \item We present A\textsuperscript{2}FM, the first adaptive agent foundation model that integrates \textit{agentic}, \textit{reasoning}, and \textit{instant} modes under a single backbone via a self-adaptive router.
    \item We propose A two-stage process: (i) supervised route-then-align fine-tuning for mode-conditioned trajectories; (ii) APO for mode selection with adaptive sampling and cost-regularized rewards.
    \item We empirically demonstrate that A\textsuperscript{2}FM achieves state-of-the-art results at the 32B scale across agentic, reasoning, and general benchmarks, with substantial reductions in token usage and computation versus mode-forcing baselines.
\end{itemize}

%% file: sec3_related.tex
\section{Related Work}
\subsection{Agent Systems and Frameworks}

Recent agent systems extend large language model (LLM) autonomy beyond text-only reasoning by introducing planning, multi-step tool use, and collaborative decision-making pipelines~\citep{zhou2023agents,zhou2024agents2,smolagents,tang2025agent,jin2025revealselfevolvingcodeagents,hu2025owl,mai2025agent}. A common design pattern involves external orchestration: the system decomposes tasks into subgoals, assigns them to role-specialized agents such as planners, researchers, coders, and verifiers, and coordinates calls to search, browsing, and code-execution tools. These frameworks demonstrate strong performance on complex, open-ended problems through explicit workflow design and tool integration~\citep{li2023camel,hong2024metagpt,qian2023chatdev,fourney2024magentic,chen2023agentverse}.

Within this paradigm, multi-agent frameworks leverage division of labor and tool diversity to improve coordination and task efficiency~\citep{zhou2023agents,zhou2024agents2,smolagents,tang2025agent,xie2025aworld,zhang2024aflow}. However, they often exhibit high computational and latency overhead due to inter-agent communication and extended reasoning traces, and generalization to new domains typically requires additional prompt or workflow engineering~\citep{zeng2023flowmind,zhang2024aflow,hu2025owl}. The backbone LLMs in these systems are not trained end-to-end for multi-turn, multi-tool protocols, and therefore rely on predefined orchestration rules rather than learned decision policies~\citep{li2025coa,zhou2023recurrentgpt,liu2023llm}.

Single- and few-agent frameworks focus on tool-centric and deep-search tasks with tighter reasoning–action loops. Representative examples include WebDancer~\citep{wu2025webdancer}, WebShaper~\citep{tao2025webshaper}, WebSailor~\citep{li2025websailor}, Asearcher\citep{gao2025asearcher}, and AFM~\citep{li2025coa}, which emphasize structured browsing, information extraction, and verification, as well as OAgents~\citep{zhu2025oagents} and DeepResearcher~\citep{zheng2025deepresearcher}, which integrate planning and tool APIs based on strong proprietary backbones. These systems enhance reliability in search and code-execution tasks and introduce structured planning procedures such as step-wise verification or retry mechanisms~\citep{pang2025browsemaster,zhang2025evolvesearch,gao2025asearcher}. Nevertheless, they are largely governed by controller heuristics and fixed procedural flows, where routing among direct answering, internal reasoning, or external tool use is manually designed rather than adaptively learned from data~\citep{xue2025simpletir,qian2025toolrl,li2025torl}.

\subsection{Adaptive Reasoning Approaches}

A growing line of work improves reasoning efficiency by letting models adaptively regulate how long and when to think, rather than relying on fixed, full-length chain-of-thought (CoT) reasoning. These methods broadly fall into two complementary categories: length-aware control, which optimizes the depth of reasoning, and capability-aware routing, which decides when extended reasoning should be invoked.

\subsubsection{Length-Aware Control}
Length-aware methods reduce redundant token consumption by training models to produce concise CoT traces when full-step reasoning is unnecessary. Two main paradigms are common:  
(1) reinforcement learning with length regularization, which introduces token-length penalties or efficiency terms to balance accuracy and brevity~\citep{arora2025training,yeo2025demystifyinglongchainofthoughtreasoning,aggarwal2025l1controllinglongreasoning}; and  
(2) supervised CoT compression, which yields variable-length CoT via post-hoc summarization or by learning compressed reasoning patterns from data~\citep{xia2025tokenskipcontrollablechainofthoughtcompression,kang2024c3otgeneratingshorterchainofthought,munkhbat2025selftrainingelicitsconcisereasoning,liu2024languagemodelslearnskip}.  
These approaches demonstrate consistent token savings while maintaining task performance.

\subsubsection{Capability-Aware Routing}
Capability-aware routing focuses on when to reason by using internal signals (e.g., uncertainty, logit margins, hidden representations) to estimate difficulty and trigger appropriate reasoning depth~\citep{chen2024teachinglargelanguagemodels,he2025selfrouteautomaticmodeswitching}. Several works further adopt reinforcement learning to learn when-to-think policies~\citep{lou2025adacot,yu2025dapo}; Bimodal Policy Optimization contrasts utilities of reasoning vs.\ non-reasoning trajectories for the same input~\citep{yang2025r4bincentivizinggeneralpurposeautothinking}; and Large Hybrid Reasoning Models route by query semantics to invoke extended reasoning when beneficial~\citep{jiang2025thinkneedlargehybridreasoning}.  
Overall, this literature primarily studies binary switching between short and long CoT and generally does not incorporate end-to-end training for adaptive tool use.

%% file: sec4_method.tex
\section{Method}

\subsection{Problem Formulation}

We consider a system that processes queries $x \in \mathcal{X}$ drawn from a task mixture $\mathcal{D}$ and produces outputs $y \in \mathcal{Y}$. For each query, a \textit{router} selects one of three execution modes:
\[
\mathcal{M} = \{\texttt{instant},\ \texttt{reasoning},\ \texttt{agentic}\}, 
\quad m \sim \pi_{\mathrm{route}}(m \mid x)
\]
where $\pi_{\mathrm{route}}$ is a routing policy over the mode set $\mathcal{M}$.

Given the selected mode $m$, the system then generates an output through a \textit{mode policy} $\pi_m$, i.e.,
\[
y \sim \pi_{m}(y \mid x)
\]
Each mode induces a distinct form of trajectory $\tau_m$: direct answers for \texttt{instant}, chain-of-thought sequences for \texttt{reasoning}, and tool-interaction traces for \texttt{agentic}. We denote the decoding function of a trajectory under mode $m$ as $f_m(x, \tau_m)$.

Let $\mathrm{Acc}(x,y) \in [0,1]$ be a task-specific accuracy metric. The expected performance of mode $m$ on input $x$ is
\[
Q_m(x) \;=\; \mathbb{E}_{\tau_m \sim \pi_m(\cdot \mid x)} 
\Big[ \mathrm{Acc}\big(x, f_m(x,\tau_m)\big) \Big]
\]

Our objective is to jointly optimize the router and mode policies such that the system maximizes expected accuracy across the task mixture:
\[
\max_{\pi_{\mathrm{route}},\,\{\pi_m,f_m\}}
\;\; \mathbb{E}_{x \sim \mathcal{D}} 
\Bigg[ \sum_{m \in \mathcal{M}} \pi_{\mathrm{route}}(m \mid x)\, Q_m(x) \Bigg]
\]

This unified formulation makes explicit the two levels of decision-making: 
(1) the router $\pi_{\mathrm{route}}$ decides which mode to activate for a given query, and 
(2) the mode-specific policy $\pi_m$ determines the quality of the generated trajectory. 
Together, they provide the foundation for both supervised alignment and reinforcement learning.

\subsection{Stage 1: Route-then-Align Fine-tuning}

\paragraph{Training Data Generation.} 
The first stage centers on supervised route-then-align training, where the model learns to classify a query into one of three modes—\texttt{instant}, \texttt{reasoning}, or \texttt{agentic}—and then generate mode-consistent trajectories. Mode routing is triggered by paired \texttt{<classification>} tags, after which the model proceeds with different behaviors:

- In the \texttt{instant} mode, the model directly outputs the final prediction within \texttt{<answer>} tags, minimizing deliberation.  

- In the \texttt{reasoning} mode, the model provides a chain-of-thought wrapped in \texttt{<reasoning>} tags, followed by the final result in \texttt{<answer>} tags.  

- In the \texttt{agentic} mode, the model interleaves high-level reasoning with external tool usage. While inspired by Agent Foundation Models~\citep{li2025coa}, our design departs significantly in how $plan$ and $summary$ are used.  

\begin{itemize}
    \item $plan$: appears only once at the beginning, decomposing the query into multiple sub-goals that can be executed in parallel.  
    \item $summary$: operates dynamically across the process, allowing the system to concurrently aggregate solved sub-tasks, terminate completed threads, and open new ones when needed.  
\end{itemize}

This explicitly parallel architecture enables simultaneous multi-tool execution, substantially improving both efficiency and effectiveness of tool usage.

\paragraph{Environment Interaction (Agentic Mode).}
To support agentic execution, the policy interacts with external resources through three MCP-style tool interfaces that return structured observations:

\begin{itemize} 
    \item $web\_search$: We interface with the Google Search engine for information retrieval, conducting web searches via the SerpAPI\footnote{\url{https://google.serper.dev/search}} tool. The core SerpAPI configuration parameters are the search query string and the number of results to return. In practice, the tool issues model-generated queries and, by default, retrieves the top 5 results for each query. Each result includes a title, a snippet, and the corresponding Uniform Resource Locator (URL). This setup provides essential support for subsequent analysis and decision-making processes. 
    
    \item $crawl\_page$: We employed a web-crawling tool authenticated via Jina API\footnote{\url{https://jina.ai/}} and summarized the retrieved content using gpt-5-mini. The tool's core configuration parameters comprise the Uniform Resource Locator (URL), the summary query. Candidate URLs are generated by the model, and the tool uses the Jina API to crawl each URL. Subsequently, we invoke the gpt-5-mini model to produce a summary for each crawled page; the summary prompt is provided in Appendix~\ref{app:summary_prompt}. 
    
    \item $code\_execute$: To ensure usability and security, we realize the code sandbox with nsjail\footnote{\url{https://github.com/google/nsjail}}, a lightweight utility that provisions isolated execution environments for Python code. By leveraging Linux namespace isolation, nsjail hardens file-system boundaries and prevents unauthorized access to host resources. A notable strength of this approach is its compatibility with containerized ecosystems (e.g., Docker), which facilitates seamless migration across diverse training and testing settings. The tool also offers fine-grained resource controls. during the training process, we cap CPU time at 5 seconds and memory at 5 GB to keep code execution strictly bounded. 
\end{itemize}

Unless otherwise noted, tokens originating from \texttt{<tool\_response>} are treated as observations and excluded from the optimization loss; the model is trained only on its own generated tokens, so learning focuses on routing, reasoning, and action selection rather than memorizing tool outputs.

The \texttt{agentic} trajectory begins with a \texttt{<plan>} section, executes $N$ tools in parallel (each wrapped in \texttt{<tool\_call>} tags), and collects their results within \texttt{<tool\_response>} tags. These tool results are masked during training—following the strategy of Search-R1~\citep{jin2025searchr1}—so that during the fine-tuning stage, the model focuses on reasoning and routing, not memorization of tool outputs. A final \texttt{<summary>} may be added before producing the \texttt{<answer>}. Details of the training data trajectories template are deferred to Appendix~\ref{app:Template}.

\begin{figure}[!ht]
    \centering
    \begin{overpic}[width=0.4\linewidth]{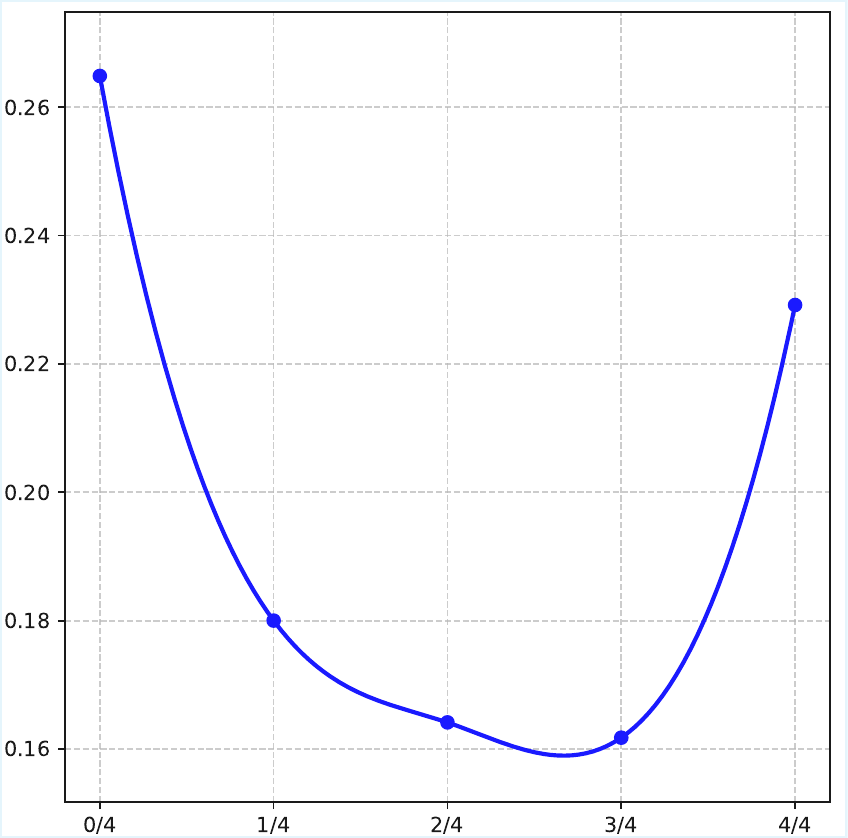}
        \put(-10,35){\rotatebox{90}{\small Sample ratio}}
        \put(40,-7){\small Easiness (x/4)}
    \end{overpic}
    \hspace{0.04\linewidth}
    \begin{overpic}[width=0.4\linewidth]{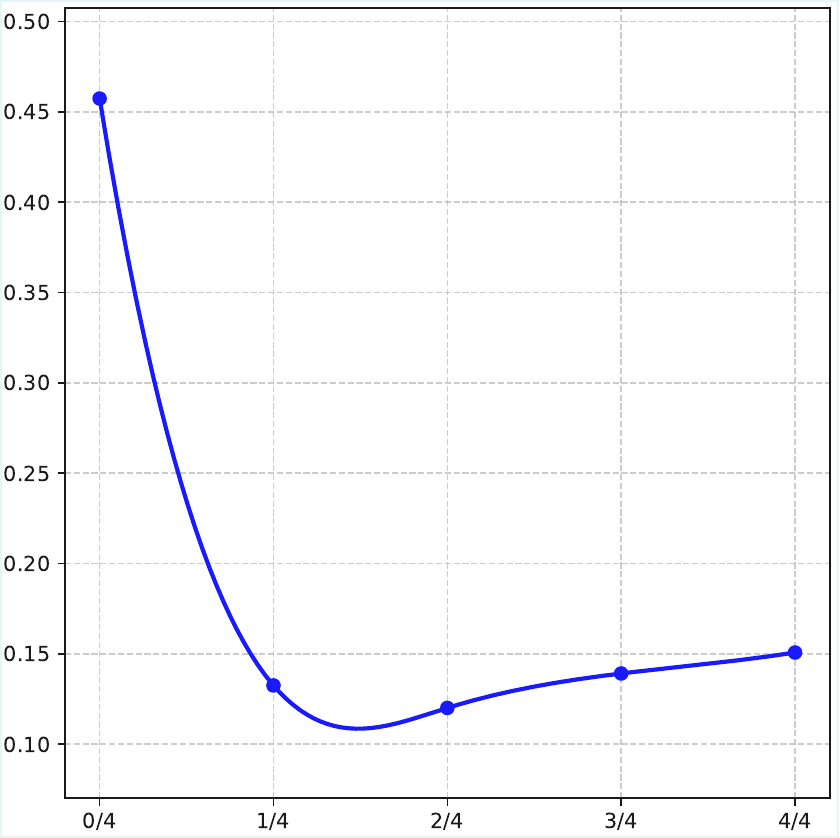}
        \put(40,-7){\small Easiness (x/4)}
    \end{overpic}

    \vspace{10pt}
    \caption{Difficulty-based sampling produces a J-shaped distribution for both the agentic (left) and reasoning (right) datasets, reducing trivially solved cases while emphasizing moderately challenging queries.}
    \label{fig:difficulty_distribution}
\end{figure}

\paragraph{Data Curation.} 
To construct diverse and challenging training data, we employ two complementary heuristics. First, a difficulty-based sampling strategy adjusts the natural distribution of task success. In the raw dataset, sampling typically yields a U-shaped distribution—tasks are either consistently solved or consistently unsolved, with few of intermediate difficulty. To enhance the representation of moderately challenging cases, we deliberately downsample “always-solved’’ instances, resulting in a J-shaped distribution that better balances task difficulty (see Fig.~\ref{fig:difficulty_distribution}). This modification ensures broader coverage across the difficulty spectrum and exposes the model to queries near its decision boundaries.  
Second, we explicitly address classification-ambiguous queries where routing decisions are uncertain. Instead of relying on majority votes, we assign the mode label corresponding to the trajectory that achieves the highest accuracy on that query. This encourages stronger supervision near ambiguous boundaries, improving the model’s ability to discriminate between reasoning, agentic, and instant behaviors when multiple modes appear plausible.  

For distillation, we adopt mode-specialized teachers: DeepSeek R1~\citep{guo2025deepseekr1} (strong reasoning capability) for the reasoning mode, and DeepSeek V3.1~\citep{deepseekai2025deepseekv31} (broad general competence) for the agentic and instant modes. This complementary setup ensures that each mode benefits from the most suitable teacher, leading to more effective supervision and consistent alignment under a unified backbone.  

\subsection{Stage2: Adaptive Policy Optimization}
\begin{figure}[t]
    \centering
    \vspace{-10pt}
    \includegraphics[width=\linewidth]{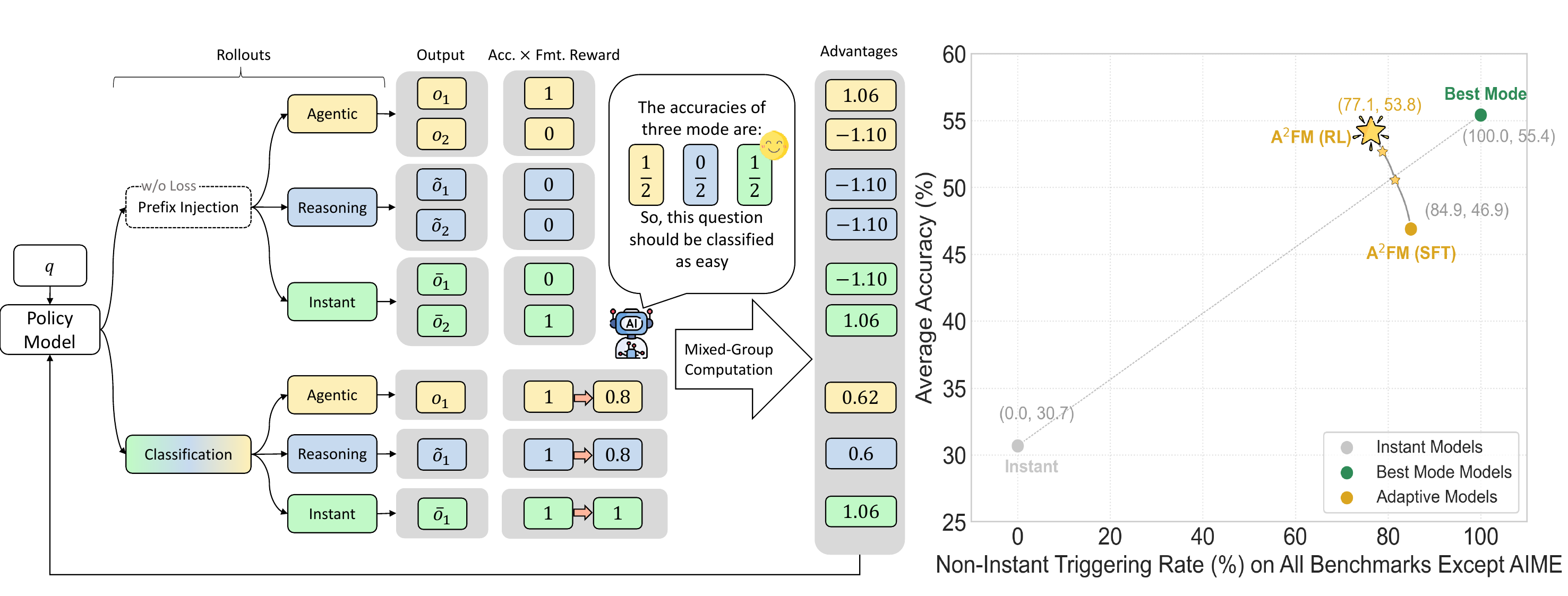}
    \vspace{-12pt}
    \caption{Overview of Adaptive Policy Optimization (APO). \textbf{Left:} Rollout and reward process. For each query, mode-specific rollouts are generated either by prefix injection (forced agentic/reasoning/instant) or by adaptive classification. Both prefix-injection tokens and tool-response tokens are excluded from loss since they are not model-generated. \textbf{Right:} Accuracy–efficiency trajectory under APO, showing how A\textsuperscript{2}FM progressively approaches the Pareto frontier by improving accuracy while reducing non-instant triggering (excluding AIME24/25).}
    \label{fig:rl}
\end{figure}

Aiming to tackle the challenges in balancing accuracy and efficiency in query processing, we propose Adaptive Policy Optimization (APO), which equips the model with dynamic mode-selection—choosing among agentic, reasoning, or instant—based on task demands.  

APO builds on GRPO~\citep{shao2024deepseekmath_grpo} but extends it in two key ways: (i) a tailored rollout strategy that ensures sufficient exploration of all modes, avoiding under-sampling; and (ii) an adaptive reward that explicitly encodes the accuracy–efficiency trade-off, rather than relying on coarse binary rewards. This design enables the model to not only maintain correctness but also minimize unnecessary computation, achieving accurate yet cost-effective routing.

{\textbf{Rollout Strategy.}}
For each query, APO performs both \textit{mode-forced rollouts} and \textit{adaptive rollouts}. In the forced setting, the model is compelled to operate in each of the three modes—agentic, reasoning, and instant—for $\rho$ rollouts per mode, where $\rho$ is a tunable hyperparameter. Mode enforcement is implemented via prefix injection, in which a pre-specified classification tag is inserted at the beginning of the model’s response. This guarantees that every query is explored under all modes, enabling an unbiased estimate of their relative success rates. Such statistics provide the foundation for adaptive rewards (See Adaptive Reward below), which encourages the use of more accurate modes while preferentially selecting the instant mode whenever possible to minimize token consumption.

The enforced prompts for each mode are outlined below:

\begin{enumerate}
    \item \texttt{Reasoning Mode}: This task requires complex logical reasoning (such as mathematical proofs, multi-step problem solving) and causal analysis, so I will select \texttt{reasoning\_agent}. \texttt{<classification>} \texttt{reasoning\_agent} \texttt{</classification>}

    \item \texttt{Agentic Mode}: This task requires acquiring real-world information (such as news and data) or executing code (such as programming problems, data processing, or statistics), so I will select \texttt{agentic\_agent}. \texttt{<classification>} \texttt{agentic\_agent} \texttt{</classification>}

    \item \texttt{Instant Mode}: This task needs no real-world info, code, or complex reasoning—just basic knowledge or brief responses, so I will select \texttt{instant\_agent}. \texttt{<classification>} \texttt{instant\_agent} \texttt{</classification>}
\end{enumerate}

In addition, APO samples $\gamma$ \textit{adaptive rollouts}, where $\gamma$ is another hyperparameter controlling the number of trials in which the model autonomously selects its operating mode. These adaptive samples provide the opportunity to reward correct self-routing, thereby reinforcing accurate and cost-efficient mode selection.

{\textbf{Accuracy Reward.}}
We adopt the LLM-as-Judge framework~\citep{zheng2023llmjudge} to assess correctness, where a judge model $M_j$ provides binary feedback:  
\[
r_{\text{acc}} = \mathbb{I}[M_j(x,\hat y)=1].
\]
This avoids the limitations of rule-based metrics (e.g., F1, EM) that cannot fully capture the validity of open-ended outputs.

{\textbf{Adaptive Reward.}}\label{para:adap_reward}
To further guide proper mode selection, we introduce an adaptive reward that explicitly favors minimal-effort solutions on easy tasks. Intuitively, if a query can already be solved by the \texttt{instant} mode, then choosing a more costly mode (reasoning or agentic) should incur a penalty. Formally, we label a query as easy if the instant mode achieves accuracy above a threshold $\tau$. For such easy queries, the reward is defined as
\[
r_{\text{adaptive}} =
\begin{cases}
1 - p^\alpha, & \text{if a non-instant mode is chosen}, \\
1, & \text{otherwise},
\end{cases}
\]
where $p$ is the empirical success rate of all forced rollouts on that query, and $\alpha > 0$ is a scaling factor. This design ensures that correct use of the instant mode always receives full reward, while reasoning or tool use on easy tasks is penalized in proportion to how confidently the query could be solved instantly. For hard queries, no penalty is applied, prioritizing the focus on correctness.

{\textbf{Format Reward.}}
We enforce strict schema compliance: if the output violates the mode-specific format (e.g., tool tags in \texttt{instant}), the reward is zero; otherwise it is one:
\[
r_{\mathrm{format}} =
\begin{cases}
1, & \text{if $y$ matches the format of mode $m$}, \\
0, & \text{otherwise}.
\end{cases}
\]

{\textbf{Total Reward.}}
The final reward combines accuracy, adaptivity, and format constraints:
\[
r_{\mathrm{total}} = r_{\mathrm{accuracy}} \times r_{\mathrm{adaptive}} \times r_{\mathrm{format}}.
\]
This multiplicative design ensures that failure in any single component (e.g., wrong answer, misuse of modes, or format violation) immediately results in reward deduction, thereby enforcing strict correctness while still encouraging efficiency and proper schema adherence.

We adopt GRPO's objective calculation with certain specializations. Specifically, we strictly employ on-policy training, as varying model capabilities may affect query classification, thereby leading to an unstable training process. Additionally, to expedite training and explore more efficient mode selections, we omit the computation of KL divergence. Consequently, given an actor parameterized by $\theta$, for each question $x_i$, APO samples a group of outputs $\{y_{i1}, y_{i2}, \cdots, y_{iG}\}$ from the old policy $\pi_{\theta_{\mathrm{old}}}$ and then optimizes the policy model by maximizing the following objective:
\small 
\begin{align*}
J_{\mathrm{APO}}(\theta) &= \mathbb{E}_{[\substack{xi \sim \mathcal{D}, \{y_{ij}\}_{j=1}^G \sim \pi_{\theta_{\mathrm{old}}}(Y|xi)}]} \\
& \frac{1}{G} \sum_{t=0}^{G}  \frac{1}{|y_{ij}| - 1} \sum_{t=0}^{|y_{ij}| - 1}
\min\left\{ 
\frac{\pi_{\theta}\left(o_t^{(ij)} \mid s_t^{(ij)}\right)}
     {\pi_{\theta_{\textit{old}}}\left(o_t^{(ij)} \mid s_t^{(ij)}\right)} \hat{A}_{ij},
\mathrm{clip}\!\left( 
\frac{\pi_{\theta}\left(o_t^{(ij)} \mid s_t^{(ij)}\right)}
     {\pi_{\theta_{\textit{old}}}\left(o_t^{(ij)} \mid s_t^{(ij)}\right)},
1-\varepsilon, 1+\varepsilon 
\right) \hat{A}_{ij}
\right\}
\end{align*}
where $o_{t}^{\left( ij \right)}$ is the token at index $t$ in response $y_{ij}$; $s_{t}^{\left( ij \right)}:=(x_i,a_{0}^{\left( ij \right)},...,a_{t-1}^{\left( ij \right)})$ is the prefix context when generating $a_{t}^{\left( ij \right)}$;$G:=3\times\rho+\gamma$ is the sum of all mode trajectory in the rollout stage.

This improved framework with rollout strategy and adaptive reward ensures more stable policy optimization and targeted mode selection. Through this refined reinforcement learning paradigm, A\textsuperscript{2}FM achieves enhanced performance in adaptive mode routing, while simultaneously boosting overall accuracy and efficiency across diverse query scenarios.

%% file: sec5_exp.tex
\section{Experiment}

\subsection{Dataset Composition}\label{app:dataset_composition}
Our dataset sources encompass a variety of task types and domains. We specifically chose not to focus on easy datasets but rather to select data from more challenging and complex sources. These include datasets such as Webdancer\citep{wu2025webdancer}, Webshaper\citep{tao2025webshaper}, Taskcraft\citep{shi2025taskcraft}, 2Wiki\citep{ho20202wiki}, Musique\cite{ho20202wiki}, and HotpotQA\citep{yang2018hotpotqa} for agentic-related tasks, and mathematic reasoning datasets like Skywork-Math\citep{zeng2024skywork}, DAPO-Math-17K\citep{yu2025dapo}, and GSM8K\citep{cobbe2021gsm8k}. Additionally, code-related data sources like Skywork-Code\citep{zeng2024skywork} and Codeforces\citep{penedo2025codeforces} were incorporated for programming tasks. Importantly, while we did not curate separate datasets solely for the instant mode, we deliberately retained relatively simple examples within the search and reasoning datasets. This ensures balance across the three modes and allows the model to naturally learn to apply the instant mode on straightforward tasks.
Finally, we created 5289, 2829, and 2890 trajectories for agentic, reasoning and instant modes respectively during the route-then-align SFT phase and 6000 question-answer pairs during the APO phase.

\subsection{Experimental Setup}\label{sec:exp_setup}
We train A\textsuperscript{2}FM on a mixture of agentic, reasoning, and code-related datasets, while also retaining relatively simple cases to enable learning of the instant mode. Full dataset composition and sampling strategy are provided in Appendix~\ref{app:dataset_composition}. 

\textbf{Baselines.}
For comparison, we conducted a comprehensive evaluation of our trained model against three categories of systems: (1) general-purpose large language models (LLMs), including closed-source models such as GPT-4.1\citep{openai2025gpt41}, o1~\citep{jaech2024openaio1}, Claude-4-Sonnet\citep{anthropic2025claude4}, and open-source models DeepSeek-R1\citep{guo2025deepseekr1}, Qwen2.5-Instruct-32B~\citep{yang2024qwen25}, Qwen3-32B\citep{yang2025qwen3}, QwQ-32B\citep{qwen2025qwq32b}; (2) state-of-the-art agent frameworks, such as OAgents\citep{zhu2025oagents}; and (3) Agent Foundation Models, including DeepDive\citep{lu2025deepdive}, WebDancer\citep{wu2025webdancer}, WebSailor\citep{li2025websailor}, Asearcher‑Web\citep{gao2025asearcher}, AFM‑Search\citep{li2025coa}, SimpleTIR\citep{xue2025simpletir}, EffectiveTIR\citep{bai2025effectivetir}, AutoTIR\citep{wei2025autotir}, and ReTool\citep{feng2025retool}.

{\textbf{Evaluation Benchmarks.}}
We evaluate our model on three key types of benchmarks. First, for agentic tasks, we use BrowseComp\citep{wei2025browsecomp}, GAIA\citep{mialon2023gaia} (we test just a 103 text-only subset extracted in \citep{wu2025webdancer}), and XBench-DeepSearch\citep{chen2025xbench}, which assess the model's ability to interact with external tools like search engines and databases for information seeking. Second, for reasoning-dependent tasks, we utilize AIME24\citep{maa2024aime24}, AIME25\citep{maa2025aime25}, and Math500\citep{hendrycks2021math}, which focus on mathematical reasoning and problem-solving without relying on external information. Finally, for general knowledge and comprehension, we use GPQA-d\citep{rein2024gpqa}, SuperGPQA\citep{du2025supergpqa}, MMLU-Pro\citep{wang2024mmlupro}, and HLE\citep{phan2025hle} (We use a 500 text-only subset defined in \citep{li2025webthinker}), which test the model's ability to synthesize information and apply broad general knowledge across a wide range of domains.

\input{tables/tab_main_results_overall}

{\textbf{Implementation Details.}}
We conduct all experiments on the Qwen2.5-32B-Instruct backbone. In the route-then-align SFT stage, we train for 3 epochs with a batch size of 256, AdamW optimizer, cosine decay learning rate schedule, and a max sequence length of 32,768. In the APO stage, we train for 2 epochs with a learning rate of 1e-6, 5 warmup steps, batch size 128, and 12 rollouts per prompt (with $\rho=3$ and $\gamma=3$) capped at 65,536 tokens. We set $\alpha=2$ for adaptive reward calculation. For inference, we set temperature 1.0, top-p 0.9, top-k 20, and a max output length of 131,072.

\subsection{Main Results}
\begin{figure*}[!ht]
    \centering
    \begin{subfigure}[t]{0.48\textwidth}
        \centering
        \includegraphics[width=\linewidth]{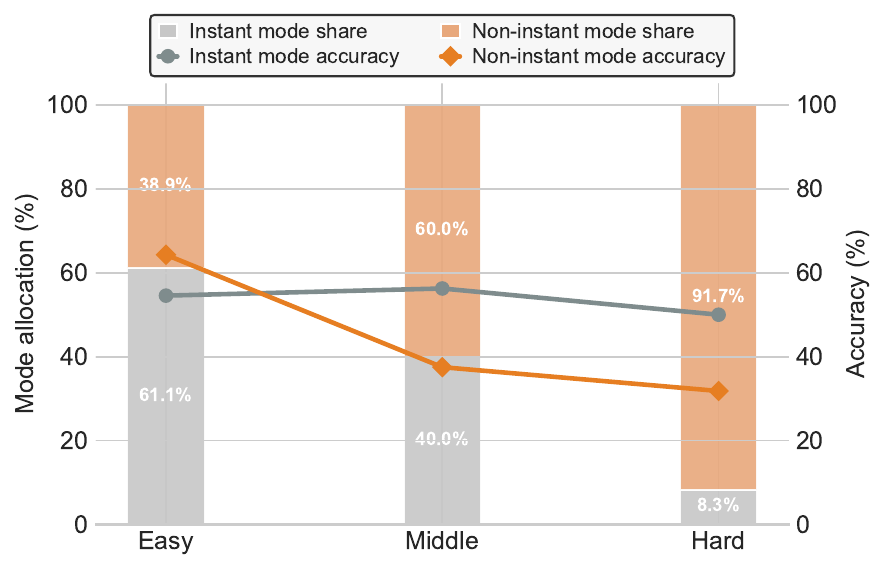}
        \caption{Relation between task difficulty, allocation ratio, and accuracy for instant and non-instant modes.}
        \label{fig:supergpqa_difficulty_modes_accuracy}
    \end{subfigure}%
    \hfill
    \begin{subfigure}[t]{0.48\textwidth}
        \centering
        \includegraphics[width=\linewidth]{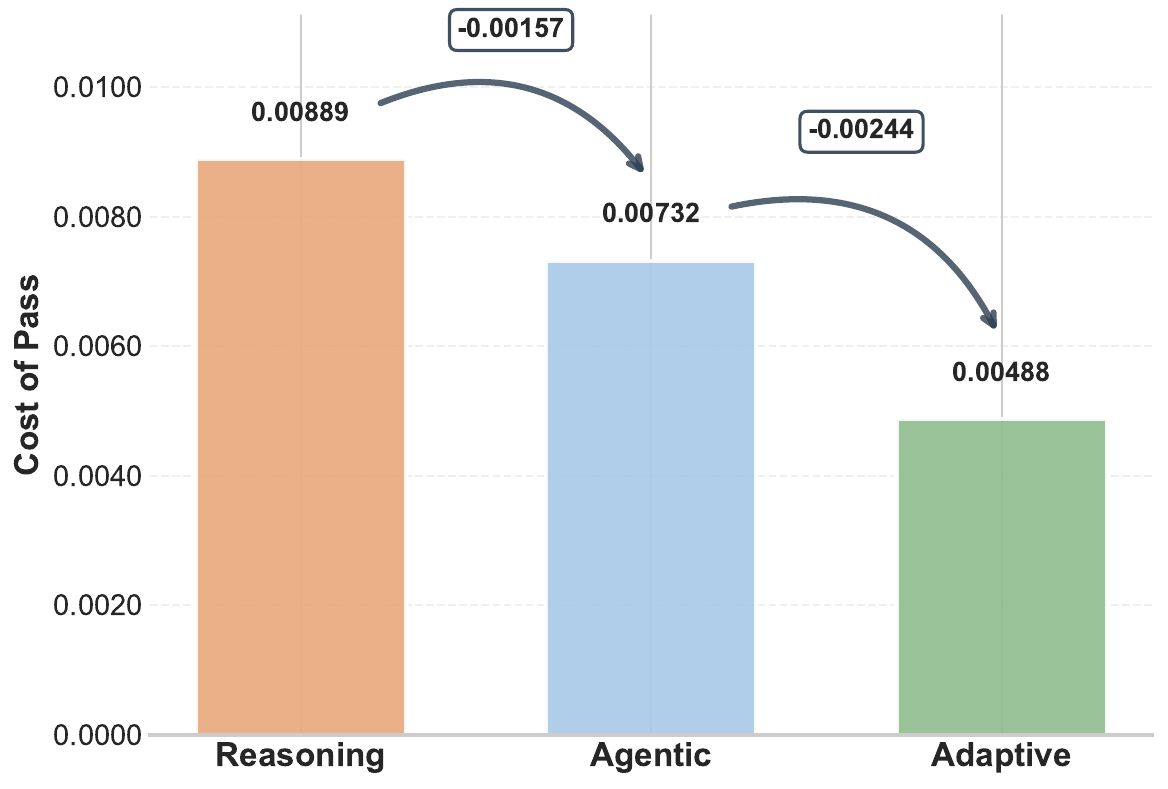}
        \caption{Cost of Pass from different modes.}
        \label{fig:token_consumption_comparison}
    \end{subfigure}
    \vspace{-8pt}
    \caption{Efficiency analysis on SuperGPQA: mode allocation vs.\ difficulty and token usage across modes.}
    \label{fig:difficulty_token_analysis}
    \vspace{-12pt}
\end{figure*}

{\textbf{Evaluation of Agentic Abilities.}}
The results in Table~\ref{tab:agentic_results} show that our adaptive mode (A\textsuperscript{2}FM) achieves consistently competitive performance across all agentic benchmarks. On XBench-DS, A\textsuperscript{2}FM reaches \textbf{56.0\%}, the highest among all baselines and surpassing the second-best model, AFM-Search (54.0\%), by +2.0 points. On GAIA, A\textsuperscript{2}FM ranks second overall, trailing only OAgents—which leverages GPT-4.1 as its backbone—while still outperforming all general-purpose LLMs (e.g., Claude-4-Sonnet at 2.6\%) and all 32B-scale agent foundation models. On BrowseComp, A\textsuperscript{2}FM achieves \textbf{12.4\%}, ranking third behind DeepDive (14.8\%) and OAgents (13.7\%), but still exceeding AFM-Search (11.1\%) and all general-purpose LLMs.

When forcing the model into the agentic mode (A\textsuperscript{2}FM-Agentic), performance further improves: on BrowseComp it rises to \textbf{14.4\%}, surpassing OAgents and achieving the second-highest score overall, just behind DeepDive (14.8\%); on GAIA it reaches \textbf{60.7\%}, establishing a new SOTA. This demonstrates that A\textsuperscript{2}FM not only delivers robust adaptive performance but also, when focused on tool-intensive reasoning, can outperform specialized frameworks in complex deep-search scenarios.

{\textbf{Evaluation of Reasoning.}}
As shown in Table~\ref{tab:reasoning_results}, A\textsuperscript{2}FM achieves reasoning performance comparable to o1 on MATH500 (\textbf{95.0\%}), while significantly outperforming all 32B-scale agent foundation models (e.g., ReTool 67.0\% on AIME24, AFM Code 59.8\% on AIME25). On AIME24, it sets a new SOTA with \textbf{74.5\%}, and on AIME25 it attains \textbf{70.4\%}, second only to o1 (79.2\%). Compared to Claude 4 Sonnet, A\textsuperscript{2}FM is stronger by +33.3 and +40.2 points on AIME24/25, underscoring reasoning capabilities far beyond even top general-purpose LLMs. Notably, adaptive and reasoning-specific variants yield identical results on AIME24/25, showing that A\textsuperscript{2}FM reliably routes nearly all AIME queries into the reasoning mode.

{\textbf{Evaluation of General Abilities.}}
The results in Table~\ref{tab:general_results} show that A\textsuperscript{2}FM delivers highly competitive general-domain performance at the 32B scale. The adaptive mode consistently outperforms instruction-tuned baselines (e.g., Qwen2.5-32B) and even surpasses the reasoning-focused QwQ. Despite starting from a 32B instruct backbone, A\textsuperscript{2}FM achieves large gains: on GPQA-d it improves by +13.6 points over Qwen2.5-32B (\textbf{63.1} vs. 49.5), and on SuperGPQA it is +15.9 points higher (\textbf{54.7} vs. 38.8), notably surpassing GPT-4.1 (50.8) and Claude 4 Sonnet (55.7).

HLE further highlights the model’s strength on integrated tasks requiring both reasoning and tool use. Here, A\textsuperscript{2}FM-Agentic achieves a remarkable \textbf{20.6\%}, exceeding the next-best baseline (QwQ, 8.2\%) by +12.4 points, while the adaptive mode (\textbf{16.7\%}) still leads all comparable 32B models by a wide margin. The gap between adaptive and agentic variants reflects that some queries are routed into the instant mode, illustrating A\textsuperscript{2}FM’s ability to balance effectiveness with efficiency even under challenging integrated settings.

{\textbf{Evaluation of Efficiency.}}
We focus on A\textsuperscript{2}FM’s efficiency by analyzing its mode allocation across task difficulty (Fig.~\ref{fig:supergpqa_difficulty_modes_accuracy}). On SuperGPQA (with human-annotated difficulty), the model uses instant mode for \textbf{61.1\%} of easy questions, but this drops to just \textbf{5.3\%} for difficult ones. This shows adaptive routing: it leans on reasoning/tool use for complex tasks and direct answers for simple ones to save computation. Notably, instant response accuracy stays stable at ~\textbf{55\%} across all difficulty levels, highlighting this mode’s robustness even for harder queries.

We further analyze efficiency through the metric \emph{cost-of-pass} (dollar cost per correct answer) on the four general benchmarks (GPQA-d, SuperGPQA, MMLU-Pro and HLE). Formally, for a model $m$ on problem $p$,
\[
\text{Cost-of-Pass}(m,p) \;=\; \frac{C_m(p)}{R_m(p)},
\]
where $R_m(p)$ is the accuracy and $C_m(p)$ is the expected inference cost, computed from input/output token counts and unit token prices. We follow the official Qwen2.5-32B-Instruct pricing, with \$0.00028 per 1k input tokens and \$0.00084 per 1k output tokens.

As illustrated in Fig.~\ref{fig:token_consumption_comparison}, The reasoning mode incurs a cost of \$0.00889 per correct answer, while the agentic mode reduces this to \$0.00732. Strikingly, the adaptive mode further lowers the cost to only \textbf{\$0.00487}, corresponding to a reduction of \textbf{45.2\%} relative to reasoning and \textbf{33.5\%} relative to agentic. In other words, adaptive execution delivers each correct answer at roughly \textbf{half the cost} of reasoning-based execution.

%% file: tables/tab_main_results_overall.tex
\begin{table*}[!ht]
  \centering
  \renewcommand{\arraystretch}{1.05}
  \setlength{\tabcolsep}{2.5pt}

  \begin{subtable}[t]{\textwidth}
    \centering
    \small
    \caption{Agentic benchmarks (GAIA, BrowseComp, XBench-DeepSearch).}
    \label{tab:agentic_results}
      \begin{tabular}{l|cc|ccc|c|ccccc}
        \toprule
        \textbf{Benchmark} &
        \textbf{\makecell{A$^2$FM}} & \textbf{\makecell{A$^2$FM\\Agentic}} &  
        \textbf{GPT-4.1} & \textbf{\makecell{Claude 4 \\ Sonnet}} &
        \textbf{\makecell{Qwen2.5 \\ Instruct-32B}} &
        \textbf{\makecell{OAgents \\ (GPT-4.1)}} & 
        \textbf{DeepDive} & \textbf{WebSailor} & \textbf{Asearcher} & \textbf{\makecell{AFM \\ Search}} \\
        \midrule
        \textbf{XBench-DS} 
        & \textbf{56.0} 
        & \underline{54.0}\textsuperscript{\scriptsize\color{red}{\,-2.0}\,\(\downarrow\)}
        & 17.0 & 21.5 & 8.7 & 47.0 & 50.5 & 53.3 & 42.1 & 54.0$^*$ \\
        \textbf{GAIA}    
        & 57.3  
        & \textbf{60.7}\textsuperscript{\scriptsize\color{teal}{\,+3.4}\,\(\uparrow\)}
        & 22.3 & 22.3$^*$ & 13.6 & \underline{58.3} & - & 53.2 & 52.8 & 55.3 \\
        \textbf{BC}   
        & 13.4  
        & \underline{14.4}\textsuperscript{\scriptsize\color{teal}{\,+1.0}\,\(\uparrow\)}
        & 1.5 & 2.6 & 0.6 & 13.7  
        & \textbf{14.8} & 10.5 & 5.2 & 11.1 \\
        \bottomrule
      \end{tabular}
  \end{subtable}
  \begin{subtable}[t]{\textwidth}
    \centering
    \small
    \caption{Reasoning benchmarks (MATH500, AIME24, AIME25).}
    \label{tab:reasoning_results}
      \begin{tabular}{l|cc|cc|c|ccccc}
        \toprule
        \textbf{Benchmark} &
        \textbf{\makecell{A$^2$FM}} & \textbf{\makecell{A$^2$FM\\Reasoning}} &
        \textbf{GPT-4.1} & \textbf{\makecell{Claude 4 \\ Sonnet}} &
        \textbf{o1} &
        \textbf{\makecell{Simple \\ TIR}} & \textbf{\makecell{Effective \\ TIR}} & \textbf{\makecell{Auto \\ TIR}} & \textbf{ReTool} & \textbf{\makecell{AFM \\ Code}} \\
        \midrule
        \textbf{MATH500} 
        & 95.0 & \underline{95.2}\textsuperscript{\scriptsize\color{teal}{\,+0.2}\,\(\uparrow\)} 
        & 92.4 & 94.0 & \textbf{96.4} & 92.9 & 86.4 & 62.6 & 93.2 & 94.6 \\
        \textbf{AIME24}  
        & \textbf{74.5} & \textbf{74.5}\textsuperscript{\scriptsize\color{teal}{\,+0.0}} 
        & 46.5 & 43.4 & \underline{74.3} & 59.9 & 42.3 & 33.3 & 67.0 & 66.7 \\
        \textbf{AIME25}  
        & \underline{70.4} & \underline{70.4}\textsuperscript{\scriptsize\color{teal}{\,+0.0}} 
        & 37.0 & 33.1 & \textbf{79.2} & 49.2 & 25.2 & 16.7 & 49.3 & 59.8 \\
        \bottomrule
      \end{tabular}
  \end{subtable}
  \begin{subtable}[t]{\textwidth}
    \centering
    \small
    \caption{General-knowledge benchmarks (GPQA-d, SuperGPQA, MMLU-Pro, HLE).}
    \label{tab:general_results}
      \begin{tabular}{l|ccc|cccc|c}
        \toprule
        \textbf{Benchmark} &
        \textbf{\makecell{A$^2$FM}} & 
        \textbf{\makecell{A$^2$FM\\Agentic}} & \textbf{\makecell{A$^2$FM\\Reasoning}} &
        \textbf{GPT-4.1} & \textbf{\makecell{Claude 4 \\ Sonnet}} &
        \textbf{\makecell{Qwen2.5 \\ Instruct-32B}} & \textbf{\makecell{Qwen3\\-32B}} &
        \textbf{\makecell{QwQ}}\\
        \midrule
        \textbf{GPQA-d}   
        & 63.1  
        & \underline{67.7}\textsuperscript{\scriptsize\color{teal}{\,+4.6}\,\(\uparrow\)}  
        & 64.7\textsuperscript{\scriptsize\color{teal}{\,+1.6}\,\(\uparrow\)} 
        & 66.3 & \textbf{68.3}   
        & 49.5 & 54.6 & 65.6 \\
        \textbf{SuperGPQA}
        & 54.7
        & \textbf{56.0}\textsuperscript{\scriptsize\color{teal}{\,+1.3}\,\(\uparrow\)}  
        & 51.2\textsuperscript{\scriptsize\color{red}{\,-3.5}\,\(\downarrow\)} 
        & 50.8 & \underline{55.7} 
        & 38.8 & 43.2 & 43.6 \\
        \textbf{MMLU-Pro} 
        & 73.8 
        & 75.8\textsuperscript{\scriptsize\color{teal}{\,+2.0}\,\(\uparrow\)} 
        & 77.0\textsuperscript{\scriptsize\color{teal}{\,+3.2}\,\(\uparrow\)} 
        & \underline{81.8} & \textbf{83.5} 
        & 69.0 & 72.7 & 76.4 \\
        \textbf{HLE}
        & \underline{16.7}  
        & \textbf{20.6}\textsuperscript{\scriptsize\color{teal}{\,+3.9}\,\(\uparrow\)}  
        & 13.4\textsuperscript{\scriptsize\color{red}{\,-3.3}\,\(\downarrow\)} 
        & 3.7 & 5.8   
        & 3.8 & 8.3 & 8.2 \\
        \bottomrule
      \end{tabular}
  \end{subtable}

  \vspace{-6pt}
  \caption{Unified results across (a) agentic, (b) reasoning, and (c) general-knowledge benchmarks. \textbf{Bold} = best; \underline{underline} = second-best. \textcolor{teal}{Teal}/\textcolor{red}{Red} superscripts indicate gain/loss of the forced mode relative to adaptive A$^2$FM. $^*$ indicates results reproduced by us. All numbers are reported as avg@1, except AIME24/25 which use avg@32.}
  \label{tab:unified_results}
  \vspace{-10pt}
\end{table*}

%% file: sec6_analysis.tex
\begin{figure}[t]
    \centering
    \includegraphics[width=\linewidth]{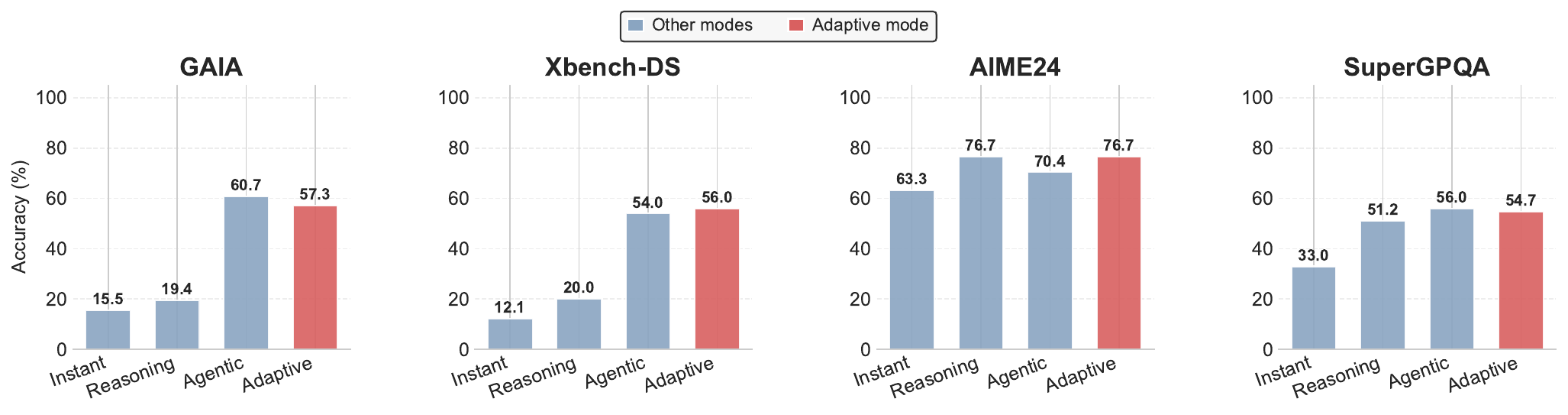}
    \vspace{-22pt}
    \caption{Comparison of adaptive mode (red) against forced single modes across four benchmarks. }
\label{fig:afm_mode_modes_compare}
\end{figure}

\section{Analysis}

\paragraph{\textbf{Approaching the Pareto Frontier under APO.}}
Figure~\ref{fig:rl} illustrates how A\textsuperscript{2}FM evolves under APO training, evaluated across all benchmarks except AIME24/25, since these tasks are routed entirely into the reasoning mode and offer no meaningful trade-off with instant responses. The three yellow stars mark checkpoints at step 10, 30, and 50, showing a clear trajectory where the adaptive model steadily reduces non-instant usage while improving accuracy. As training progresses, A\textsuperscript{2}FM moves closer to the Pareto frontier of accuracy–efficiency trade-offs, indicating that APO effectively teaches the router to allocate more queries to the instant mode without sacrificing correctness.

At convergence, A\textsuperscript{2}FM (RL) achieves an average accuracy of \textbf{53.8\%} with a non-instant ratio of \textbf{77.1\%}, compared to the “Best Mode” oracle (accuracy 55.4\%, non-instant ratio 100\%). Although the accuracy gap is marginal (–1.6 points), our adaptive model lowers non-instant triggering by \textbf{22.9} points, demonstrating that APO attains near-oracle accuracy while substantially improving efficiency.

\paragraph{\textbf{Accuracy of Mode Routing.}}
\input{tables/tab_mode_acc}

We evaluate routing accuracy on three representative datasets: GAIA (mixed agentic and reasoning), BrowseComp (agentic-dominated), and AIME24 (reasoning-dominated). For GAIA, ground-truth labels are obtained via majority voting from five strong baselines—DeepSeek-R1, DeepSeek-V3.1, GPT-4.1, GPT-5, and Claude-4-Sonnet—followed by manual verification, while BrowseComp and AIME24 are directly labeled as agentic and reasoning, respectively. As shown in Table~\ref{tab:mode_dist_acc}, our model achieves \textbf{92.2\%} accuracy on GAIA and \textbf{94.0\%} on BrowseComp. Errors on GAIA often stem from queries where both reasoning and agentic are plausible (e.g., solving math via code or explicit reasoning, see Case Study in Appendix~\ref{app:case_study}), while BrowseComp errors arise when the model overestimates its internal knowledge and avoids tool calls. On AIME24, the model reaches \textbf{100\%}, confirming robust mode discrimination on reasoning tasks.



\paragraph{\textbf{Robustness of Adaptive Routing on Complex Tasks.}}
We compare adaptive routing with forced single-mode execution (agentic, reasoning, and near-instant) across four benchmarks: GAIA, XBench-DeepSearch, AIME24, and SuperGPQA. GAIA and XBench-DeepSearch are primarily agentic-oriented with some reasoning and instant cases;  SuperGPQA contain a mix of tasks where tool use and fast responses are critical; AIME24 is reasoning-dominated. Results (Fig.~\ref{fig:afm_mode_modes_compare}) show that adaptive mode achieves competitive or superior performance in most settings. On GAIA, adaptive slightly underperforms pure agentic (57.3 vs.\ 60.7), primarily due to classification noise. On XBench-DeepSearch and AIME24, adaptive surpasses single modes, demonstrating its strength in routing composite, multi-facet queries to the appropriate execution mode rather than over-committing to a single behavior. On AIME24, adaptive mode strictly matches reasoning mode, since all queries are routed into reasoning where accuracy dominates. Overall, adaptive mode provides robust performance while retaining flexibility across diverse task types.

%% file: tables/tab_mode_acc.tex
\begin{wraptable}{r}{0.40\linewidth} 
\centering
\small
\setlength{\tabcolsep}{2.5pt}
\renewcommand{\arraystretch}{0.9}
\caption{Human-labeled mode distribution  and model classification accuracy.}
\vspace{-8pt}
\begin{tabular}{lcccc}
\toprule
Dataset & T & A & I & Acc. \\
\midrule
GAIA (N=103)        & 13.6 & 84.5 & 1.9  & 92.2 \\
BC (N=1266) & 0.0  & 100.0& 0.0  & 94.0 \\
AIME (N=30)         & 100.0& 0.0  & 0.0  & 100.0 \\
\bottomrule
\end{tabular}
\label{tab:mode_dist_acc}
\end{wraptable}

%% file: sec7_conclusion.tex
\section{Conclusion}
We present A\textsuperscript{2}FM, an Adaptive Agent Foundation Model that unifies agentic, reasoning, and instant modes under a single backbone with a self-adaptive router. By combining route-then-align training with Adaptive Policy Optimization, A\textsuperscript{2}FM achieves both high accuracy and efficiency, matching or surpassing strong baselines across agentic, reasoning, and general benchmarks. Our results highlight the promise of adaptive multi-mode modeling as a scalable path toward efficient and versatile LLM agents.

%% file: ack.tex
\newpage
\section{Contributions}

\textbf{Core Contributors}
\begin{tasks}[
            style=itemize,
            label-width=1em,
            column-sep=3.5em,
            before-skip=1.5ex,
            after-item-skip=1.5ex,
            ](2)
    \task Qianben Chen
    \task Jingyi Cao
    \task Jiayu Zhang
\end{tasks}

\textbf{Contributors}
\begin{tasks}[
            style=itemize,
            label-width=1em,
            column-sep=3.5em,
            before-skip=1.5ex,
            after-item-skip=1.5ex 
            ](2)
    \task Tianrui Qin
    \task Xiaowan Li
    \task King Zhu
    \task Dingfeng Shi
    \task He Zhu
    \task Minghao Liu
    \task Xiaobo Liang
    \task Xin Gui
    \task Ge Zhang
    \task Jian Yang
    \task Yuchen Eleanor Jiang
\end{tasks}

\textbf{Corresponding Authors}
\begin{tasks}[
            style=itemize,
            label-width=1em,
            column-sep=3.5em,
            before-skip=1.5ex,
            after-item-skip=1.5ex 
            ](2)
    \task Wangchunshu Zhou
\end{tasks}


%% file: sec9_app.tex
\section{Prompts}
\subsection{System Prompt}\label{app:system_prompt}
\setlist[enumerate]{leftmargin=1em}
\setlist[itemize]{leftmargin=1em}
\begin{tcolorbox}[
  colback=white!5!white,
  colframe=black!50!green!60!,
  colbacktitle=black!65!green!70!,
  title=\textsc{System prompt of A$^2$FM},
  fontupper=\small,
  boxrule=0.5pt,
  breakable,
]
You are required to solve the task by using one of the three agent options: agentic\_agent, reasoning\_agent, and instant\_agent.
\begin{enumerate}
    \item \textbf{Agent Options}:
\begin{enumerate}
    \item agentic\_agent: choose this agent if the task needs to \textbf{search} and \textbf{crawl} real-world / factual information (such as news and data) or \textbf{executing code} (such as programming tasks, data processing or statistics).
    \item reasoning\_agent: choose this agent if the task requires complex logical \textbf{reasoning} (such as mathematical proofs, multi-step problem solving) and causal analysis.
    \item instant\_agent: use this agent for simple tasks needing no real-world info, code, or complex reasoning. Instead, just basic knowledge or brief responses.
\end{enumerate}
    
    \item \textbf{Trajectory Formulation}:
    \begin{enumerate}
        \item You should first predict one of the three agents above within the function \texttt{<classification> ... </classification>}.
        \item Then you should formulate your thinking and processing trajectory according to the rule of the agent you choose:
        \begin{enumerate}
            \item \textbf{agentic\_agent rule}:
            \begin{enumerate}
                \item Objective:
                \begin{itemize}
                    \item Your core goal is to systematically solve user-assigned tasks by:
                    \begin{itemize}
                        \item Decomposing the task into clear goals \& paths.
                        \item Executing tools purposefully and efficiently.
                        \item Advancing all goals in parallel, while keeping each goal’s paths sequential.
                        \item Tracking progress with summaries.
                        \item Delivering a final confirmed answer only when all goals are resolved.
                    \end{itemize}
                \end{itemize}
                \item Execution Requirements:
                \begin{itemize}
                    \item Follow a logical order of functions/tools.
                    \item Parallelize independent goals; within each goal, execute paths sequentially as fallbacks.
                    \item Each step must include:
                    \begin{itemize}
                        \item thinking (before you execute tools, why this tool/path is chosen).
                        \item \texttt{<tool\_call>} execution (with correct parameters).
                        \item Use results from observations to refine next actions.
                        \item Ensure no redundant tool calls (don’t repeat identical queries).
                        \item Never assume a goal is completed without explicit verification.
                        \item Continue advancing all goals until they are resolved.
                    \end{itemize}
                \end{itemize}
                \item Functions:
                \begin{itemize}
                    \item \texttt{<plan>} Function:
                    \begin{itemize}
                        \item Role: Decompose the original task into goals and execution paths.
                        \item Rules:
                        
                        - 1–5 parallelizable goals.
                        
                        - Each goal has 1–5 paths, executed sequentially as fallback options.
                        
                        - Define success criteria for each path.
                        \item Timing: Only the first step.
                        \item Format Example:
\begin{verbatim}
<plan>
## Goal 1: [Name]
- Path 1.1: [Approach]  
- Success: [Criteria]
- Path 1.2: [Approach]  
- Success: [Criteria]
## Goal 2: [Name]
- Path 2.1: [Approach]  
- Success: [Criteria]
</plan>
\end{verbatim}
                    \end{itemize}
                    
                    \item \texttt{<summary>} Function:
                    \begin{itemize}
                        \item Role: Recap execution status and decide next actions.
                        \item Content:
                        - Plan summary (original goals/paths).
                        
                        - Execution status for each goal: Completed / In Progress / Blocked.
                        
                        - Path analysis (which worked, which failed).
                        
                        - Next steps: specify which sub-paths to run in parallel.

                        \item Timing: Every several steps, occurs when there are enough actions to summarize.
                        \item Example:
\begin{verbatim}
<summary>
## Plan Summary
[Brief recap of goals]
## Execution Status
### Goal 1: [Status]
- Path Analysis: [...]
### Goal 2: [Status]
- Path Analysis: [...]
## Next Parallel Sub-Paths
- Goal 1: Path 1.2
- Goal 2: Path 2.1
</summary>
\end{verbatim}
                    \end{itemize}
                    
                    \item \texttt{<tool\_call>} Tool:
                    \begin{itemize}
                        \item Role: Execute tools to advance goals.
                        
                        - web\_search: it has only one parameter: query (search statement). Example: \verb|{'id': xxx, 'name': 'web_search', 'arguments': {'query': 'xxx'}}|
                        
                        - crawl\_page: it has two parameters: url (valid link) and query (info to extract). Example: \verb|{'id': xxx, 'name': 'crawl_page', 'arguments': {'url': 'xxx', 'query': 'xxx'}}|
                        
                        - code\_execute: it has only one parameter: code (Markdown snippet). Example: \verb|{'id': xxx, 'name': 'code_execute', 'arguments': {'code': 'xxx'}}|

                        \item Rules:

                        - Use \textbf{1–10} tools per step (each targeting a distinct task part).
                        
                        - Each tool call must have complete, valid parameters.
                        
                        - Always prefer verifying accuracy with crawl\_page after web\_search.

                        \item Timing: All steps except \texttt{<plan>}, \texttt{<summary>}, and \texttt{<answer>}.
                    \end{itemize}
                    
                    \item \texttt{<answer>} Function:
                    \begin{itemize}
                        \item Role: Deliver the final confirmed answer.
                        \item Rules:
                        
                        - Only after all goals are resolved.
                        
                        - Must consolidate results across all goals.
                        
                        - Answer language must match task language.

                        \item Format Example:
\begin{verbatim}
<answer>
[Final Answer Content]
</answer>
\end{verbatim}
                    \end{itemize}
                \end{itemize}
                
                \item Execution Rules (Critical):
                \begin{itemize}
                    \item Parallel Goals, Sequential Paths
                    \item No Early Termination
                    \item Result Verification
                    \item Parallel Functions with Limited workers
                    \item Final Answer Condition
                \end{itemize}
            \end{enumerate}
            
            \item \textbf{reasoning\_agent rule}:
            \begin{enumerate}
                \item Trajectory:
                \begin{itemize}
                    \item Reasoning Phase: Output \texttt{<reasoning>...</reasoning>} with detailed steps (>1000 words).
                    \item Answer Phase: Present the final conclusion within \texttt{<answer>...</answer>}.
                \end{itemize}
                \item Detailed Function Specifications:
                \begin{itemize}
                    \item \texttt{<reasoning>} Function
                    \item \texttt{<answer>} Function
                \end{itemize}
                \item Notes:
                \begin{itemize}
                    \item Do not return any other functions or tools.
                    \item Output sequence is always reasoning then answer.
                    \item Reasoning must exceed 1000 words.
                \end{itemize}
            \end{enumerate}
            
            \item \textbf{instant\_agent Specification}:
            \begin{enumerate}
                \item Objective:
                \begin{itemize}
                    \item Rapidly solve tasks without tool usage or complex reasoning.
                    \item Provide clear and relevant answers.
                \end{itemize}
                \item Detailed Function Spec:
                \begin{itemize}
                    \item \texttt{<answer>} Function only.
                    \item Executed immediately, no planning or tool-calling.
                \end{itemize}
                \item Notes:
                \begin{itemize}
                    \item Must not return other functions or tools.
                    \item Entire trajectory under 300 words.
                \end{itemize}
            \end{enumerate}
        \end{enumerate}
    \end{enumerate}
    
    \item \textbf{Important Tips}:
    \begin{enumerate}
        \item You should obey the rule of the agent option you choose.
        \item Do not give an answer easily unless you are absolutely sure. The answer should be as concise as possible and avoid detailed descriptions. For example, \texttt{<answer>Beijing</answer>}.
    \end{enumerate}
\end{enumerate}
\end{tcolorbox}

\subsection{Summary Prompt}\label{app:summary_prompt}
\begin{tcolorbox}[
  colback=white!5!white,
  colframe=black!50!green!60!,
  colbacktitle=black!65!green!70!,
  title=\textsc{Case Trajectory of Post-RL model on same question above.},
  fontupper=\small,
  boxrule=0.5pt,
  breakable,
]
Target: Extract all content from a web page that matches a specific web search query, ensuring completeness and relevance. (No response/analysis required.)
        
web search query: 
...

Clues and ideas: 
...
        
Searched Web Page: 
...

Important Notes:\\
- Summarize all content (text, tables, lists, code blocks) into concise points that directly address the query and clues, and ideas.\\
- Preserve and list all relevant links ([text](url)) from the web page.\\
- Summarize in three points: web search query-related information, clues and ideas-related information, and relevant links with descriptions.\\
- If no relevant information exists, just output "No relevant information."
\end{tcolorbox}

\section{Case Study}\label{app:case_study}

\begin{tcolorbox}[
  colback=white!5!white,
  colframe=black!50!green!60!,
  colbacktitle=black!65!green!70!,
  title=\textsc{Case Trajectory of Post-SFT model on easy question.},
  fontupper=\small,
  boxrule=0.5pt,
  breakable,
]
\input{cases/bert_before}
\end{tcolorbox}

\begin{tcolorbox}[
  colback=white!5!white,
  colframe=black!50!green!60!,
  colbacktitle=black!65!green!70!,
  title=\textsc{Case Trajectory of Post-RL model on same question above.},
  fontupper=\small,
  boxrule=0.5pt,
  breakable,
]
\input{cases/bert_after}
\end{tcolorbox}

\clearpage
\section{Template}\label{app:Template}
\input{tables/tab_templates}

%% file: cases/bert_before.tex
{\color{blue!30!black}
\textit{Question}: How many more blocks (also denoted as layers) in BERT base encoder than the encoder from the architecture proposed in Attention is All You Need?
}

\setlist[enumerate,1]{label=Goal \arabic*:, leftmargin=*, labelindent=1em, itemsep=10pt}  
\setlist[enumerate,2]{label=Path \arabic{enumi}.\arabic*:, leftmargin=2.5em, itemsep=5pt} 
\setlist[itemize,1]{leftmargin=*, label=-}  

\begin{tcolorbox}[
    colback=orange!90!black!80!,  
    colframe=orange!90!black!80!, 
    boxrule=0.5pt,                
    center,                       
    fontupper=\footnotesize\bfseries, 
    halign=center,                
    valign=center,                
    colupper=white!90!,               
    height=5mm,                   
    left=2pt, right=2pt
]
Classification
\end{tcolorbox}
{\color{red!60!black}

This question asks for a specific architectural comparison (number of blocks/layers in BERT Base vs. the Transformer in 'Attention is All You Need'), which requires precise factual lookup about model definitions. Because it needs accurate, up-to-date model parameter information rather than simple common sense, the toolcalling\_agent — which can query external sources — is the appropriate choice.

<classification> \\
\textbf{toolcalling\_agent} \\
</classification>
}

\begin{tcolorbox}[
    colback=orange!90!black!80!,  
    colframe=orange!90!black!80!, 
    boxrule=0.5pt,                
    center,                       
    fontupper=\footnotesize\bfseries, 
    halign=center,                
    valign=center,                
    colupper=white!90!,               
    height=5mm,                   
    left=2pt, right=2pt
]
First tool call round
\end{tcolorbox}
{\color{black}
Now it's step 1, I need to make a plan.

<plan>
\begin{enumerate}
    \item Determine the number of encoder blocks (layers) in BERT Base
    \begin{enumerate}
        \item Search authoritative BERT paper and official codebase
        \begin{itemize}
            \item Core approach: Use web\_search to locate the BERT paper (Devlin et al., 2018) and the Hugging Face or official BERT model repository, then crawl\_page to extract the explicit encoder-layer count from the paper’s experimental setup or model configuration file.
            \item Success: A direct citation or model config line (e.g., “num\_hidden\_layers: X”) explicitly stating the encoder block count for BERT Base is extracted and saved.
        \end{itemize}
        
        \item Query popular NLP model resources and documentation
        \begin{itemize}
            \item Core approach: Use web\_search to find Hugging Face model card, PyTorch/Transformer documentation, or TensorFlow model card for BERT Base, then crawl\_page to extract the stated number of encoder blocks.
            \item Success: At least one high-quality model resource (model card or documentation) explicitly lists the encoder block count for BERT Base.
        \end{itemize}
        
        \item Inspect canonical model configuration code (fallback)
        \begin{itemize}
            \item Core approach: Use web\_search to locate the canonical configuration file or constructor in a framework (e.g., bert-base-uncased) and crawl\_page or code\_execute to read the model config and extract the layer count programmatically.
            \item Success: The layer count is programmatically read from the model config and matches the value from Path 1.1 or 1.2.
        \end{itemize}
    \end{enumerate}

    \item Determine the number of encoder blocks (layers) in the Transformer architecture from "Attention is All You Need"
    \begin{enumerate}
        \item Locate and extract from the original paper
        \begin{itemize}
            \item Core approach: Use web\_search to find the original paper (Vaswani et al., 2017) and crawl\_page to extract the encoder layer count from the experimental setup or model configuration section (e.g., Table/sections describing the base model).
            \item Success: A direct citation or explicit statement from the paper that gives the encoder block count for the base Transformer model is extracted.
        \end{itemize}
        
        \item Consult authoritative implementations and model cards
        \begin{itemize}
            \item Core approach: Use web\_search to find official or widely-used implementations (e.g., TensorFlow official models, Hugging Face “transformer-base” if available, or the original repo) and crawl\_page to extract the encoder block count.
            \item Success: At least one authoritative implementation explicitly lists the encoder block count for the base Transformer model from the paper.
        \end{itemize}
        
        \item Verify via secondary literature or reviews
        \begin{itemize}
            \item Core approach: Use web\_search to find reviews, tutorials, or survey papers that summarize the architecture of “Attention is All You Need” and extract the encoder layer count.
            \item Success: Secondary sources corroborate the encoder block count from the primary paper and provide consistent numbers.
        \end{itemize}
    \end{enumerate}

    \item Compute the difference (BERT Base encoder blocks minus Transformer base encoder blocks) and verify accuracy
    \begin{enumerate}
        \item Direct numeric subtraction and cross-check
        \begin{itemize}
            \item Core approach: After Goals 1 and 2 return values, compute difference = (BERT Base encoder blocks) - (Transformer base encoder blocks). Verify consistency across at least two independent sources per model.
            \item Success: The computed difference is produced and matches across at least two independent sources for each model.
        \end{itemize}
        
        \item Programmatic verification using model configs
        \begin{itemize}
            \item Core approach: Use code\_execute to load model configs (or synthetic config files) for BERT Base and the base Transformer and programmatically extract the encoder layer counts, then compute the difference.
            \item Success: Programmatic extraction yields the same layer counts and the computed difference equals the result from Path 3.1.
        \end{itemize}
        
        \item Cite final difference with provenance
        \begin{itemize}
            \item Core approach: Prepare a concise statement of the difference and list the provenance (paper URLs, model card links) that support the numbers used.
            \item Success: A short, unambiguous answer stating how many more blocks are in BERT Base than in the base Transformer, with at least two cited sources (one per model).
        \end{itemize}
    \end{enumerate}
\end{enumerate}

</plan>
}

Now it's step 2, I need to make tool\_calls. I will advance both goals in parallel. For Goal 1 (BERT Base encoder blocks) the primary path is to locate the BERT paper and its model config; for Goal 2 (Transformer base encoder blocks) the primary path is to locate the 'Attention is All You Need' paper and its model config. I'll perform web searches for each paper/model in parallel to find authoritative sources that state the encoder layer counts. After receiving those search results I'll crawl the specific pages to extract the explicit number of encoder blocks/layers for each model.

{\color{blue!60!black}
<tool\_call>

    \{"name": "web\_search", "arguments": \{"query": "BERT: Pre-training of Deep Bidirectional Transformers for Language Understanding number of encoder layers"\}\}
    
    \{"name": "web\_search", "arguments": \{"query": "Attention Is All You Need number of encoder layers base model"\}\}
    
</tool\_call>
}
{\color{gray!60!black}

<tool\_response>

\begin{tcolorbox}[title=Results for Tool Call 1]
    \textbf{Query:} ``BERT: Pre-training of Deep Bidirectional Transformers for Language Understanding number of encoder layers'' \\
    \textbf{Total Results:} 5 \\
    \vspace{0.3em} 
    
    \begin{enumerate}[label=\arabic*., leftmargin=*, itemsep=0.5em]
        \item \textbf{Source:} \href{https://arxiv.org/abs/1810.04805}{arXiv:1810.04805} (J. Devlin et al., 2018, Cited by 143,290) \\
              \textit{Key Content:} Mentions BERT as a pre-trained deep bidirectional model, with fine-tuning via an additional output layer (encoder layer count not explicitly stated here).
        
        \item \textbf{Source:} \href{https://aclanthology.org/N19-1423.pdf}{ACL Anthology: N19-1423} (J. Devlin et al., 2019, Cited by 143,290) \\
              \textit{Key Content:} Similar to the arXiv version; focuses on fine-tuning performance, no explicit encoder layer count.
        
        \item \textbf{Source:} \href{https://medium.com/@varunsivamani/bert-bidirectional-encoder-representations-from-transformers-4897ec2e9a06}{Medium Article} \\
              \textit{Key Content:} Explicitly states: \textbf{BERT-Base has 12 transformer layers}, 12 attention heads, and 110 million parameters; BERT-Large has 24 transformer layers.
        
        \item \textbf{Source:} \href{https://www.semanticscholar.org/paper/BERT%3A-Pre-training-of-Deep-Bidirectional-for-Devlin-Chang/df2b0e26d0599ce3e70df8a9da02e51594e0e992}{Semantic Scholar} \\
              \textit{Key Content:} Summarizes BERT's pre-training of bidirectional representations, no explicit encoder layer count.
        
        \item \textbf{Source:} \href{https://blog.paperspace.com/bert-pre-training-of-deep-bidirectional-transformers-for-language-understanding/}{Paperspace Blog} \\
              \textit{Key Content:} Introduces BERT's improvement on fine-tuning techniques, no explicit encoder layer count.
    \end{enumerate}
\end{tcolorbox}

\begin{tcolorbox}[title=Results for Tool Call 2]
    \textbf{Query:} ``Attention Is All You Need number of encoder layers base model'' \\
    \textbf{Total Results:} 5 \\
    \vspace{0.3em}
    
    \begin{enumerate}[label=\arabic*., leftmargin=*, itemsep=0.5em]
        \item \textbf{Source:} \href{https://papers.neurips.cc/paper/7181-attention-is-all-you-need.pdf}{NeurIPS 2017 Paper} (A. Vaswani et al., Cited by 192,399) \\
              \textit{Key Content:} Explicitly states: \textbf{The encoder is composed of a stack of }$N=6$\textbf{ identical layers} (each with multi-head self-attention and feed-forward sublayers).
        
        \item \textbf{Source:} \href{https://arxiv.org/abs/1706.03762}{arXiv:1706.03762} (Jun 12, 2017) \\
              \textit{Key Content:} Introduces the Transformer as a replacement for recurrent/convolutional models, no explicit encoder layer count.
        
        \item \textbf{Source:} \href{https://en.wikipedia.org/wiki/Attention_Is_All_You_Need}{Wikipedia} \\
              \textit{Key Content:} Incorrectly mentions ``encoder and decoder were both 8 layers of bidirectional LSTM'' (misrepresents Transformer's structure; original paper uses 6 encoder layers).
        
        \item \textbf{Source:} \href{https://medium.com/data-science-collective/attention-is-all-you-need-661cb8db5f21}{Medium Article} (3 months ago) \\
              \textit{Key Content:} Confirms the original paper's architecture: \textbf{The encoder stack contains 6 identical layers}, and the decoder stack also contains 6 layers.
        
        \item \textbf{Source:} \href{https://ai.plainenglish.io/i-finally-understood-attention-is-all-you-need-after-so-long-heres-how-i-did-it-263b46273f9f}{Plain English AI Blog} (Jul 12, 2025) \\
              \textit{Key Content:} States: \textbf{The encoder has 6 layers}, each consisting of 2 sublayers (multi-head self-attention + feed-forward network).
    \end{enumerate}
\end{tcolorbox}

</tool\_response>
}

\begin{tcolorbox}[
    colback=orange!90!black!80!,  
    colframe=orange!90!black!80!, 
    boxrule=0.5pt,                
    center,                       
    fontupper=\footnotesize\bfseries, 
    halign=center,                
    valign=center,                
    colupper=white!90!,               
    height=5mm,                   
    left=2pt, right=2pt
]
Second tool call round
\end{tcolorbox}

I will advance both goals in parallel using the primary path for each: verify the encoder layer counts directly from the original papers (BERT and 'Attention Is All You Need') by crawling the PDFs found in the web\_search results. I'll extract the explicit statements for BERT Base's encoder blocks and for the base Transformer's encoder layers, then compute the difference. If the PDFs don't contain the explicit counts, I'll use fallback paths (model cards or official repos) in subsequent steps, but first I'll attempt direct extraction from the paper PDFs.
{\color{blue!60!black}
<tool\_call>

\{"name":"crawl\_page", "arguments": \{"url": "https://arxiv.org/pdf/1810.04805.pdf", "query": "number of encoder layers BERT Base, number of hidden layers, BERT-base encoder blocks"\}\}

\{"name":"crawl\_page", "arguments": \{"url": "https://papers.neurips.cc/paper/7181-attention-is-all-you-need.pdf", "query": "Encoder: The encoder is composed of a stack of N = 6 identical layers"\}\}

</tool\_call>
}

{\color{gray!60!black}
<tool\_response>

\begin{tcolorbox}[title=Results for Tool Call 1]
    \textbf{Page 1} \href{https://arxiv.org/pdf/1810.04805.pdf}{[https://arxiv.org/pdf/1810.04805.pdf]} \textbf{Summary:}
    \begin{itemize}[leftmargin=*, itemsep=0.4em, label=-]
        \item \textbf{Model architecture:} BERT is a multi-layer bidirectional Transformer encoder. The paper denotes the number of layers (i.e., Transformer blocks) as $L$ and the hidden size as $H$.
        
        \item \textbf{BERTBASE (commonly “BERT‑Base”) specifications:}
              \begin{itemize}[leftmargin=1.5em, itemsep=0.2em, label=-]
                  \item Number of encoder layers / Transformer blocks ($L$): 12
                  \item Hidden size ($H$): 768
                  \item Number of self‑attention heads ($A$): 12
                  \item Feed‑forward/filter size: $4H = 3072$
                  \item Total parameters: ~110M
              \end{itemize}
        
        \item \textbf{(For context) BERTLARGE:} $L = 24$, $H = 1024$, $A = 16$, total parameters $\approx$ 340M.
        
        \item \textbf{Note:} “layers” in the paper explicitly refers to Transformer encoder blocks (i.e., encoder layers).
    \end{itemize}
\end{tcolorbox}

\begin{tcolorbox}[title=Results for Tool Call 2]
    \textbf{Page 1} \href{https://papers.neurips.cc/paper/7181-attention-is-all-you-need.pdf}{[https://papers.neurips.cc/paper/7181-attention-is-all-you-need.pdf]} \textbf{Summary:}
    \begin{itemize}[leftmargin=*, itemsep=0.4em, label=-]
        \item \textbf{Exact statement:} "Encoder: The encoder is composed of a stack of $N = 6$ identical layers." (Section 3.1)
        
        \item \textbf{Encoder layer structure:}
              \begin{itemize}[leftmargin=1.5em, itemsep=0.2em, label=-]
                  \item Each of the $N = 6$ layers has two sub-layers:
                        \begin{enumerate}[leftmargin=1.2em, itemsep=0.1em, label=1., align=left]
                            \item Multi-head self-attention mechanism.
                            \item Position-wise fully connected feed‑forward network.
                        \end{enumerate}
                  \item Residual connections are applied around each sub-layer, followed by layer normalization:
                        \begin{itemize}[leftmargin=1.2em, itemsep=0.1em, label=-]
                            \item Output form: $\text{LayerNorm}(x + \text{Sublayer}(x))$.
                        \end{itemize}
                  \item All sub-layers and embedding layers produce outputs of dimension $d_{\text{model}} = 512$ to facilitate residual connections.
              \end{itemize}
        
        \item \textbf{Feed‑forward sub-layer details (applies to each encoder layer):}
              \begin{itemize}[leftmargin=1.5em, itemsep=0.2em, label=-]
                  \item FFN formula: $\text{FFN}(x) = \max(0, x W_1 + b_1) W_2 + b_2$
                  \item Input/output dimensionality: $d_{\text{model}} = 512$
                  \item Inner layer dimensionality: $d_{\text{ff}} = 2048$
                  \item Can be seen as two $1 \times 1$ convolutions (position-wise, applied identically at every position).
              \end{itemize}
        
        \item \textbf{Attention used in encoder:}
              \begin{itemize}[leftmargin=1.5em, itemsep=0.2em, label=-]
                  \item Multi-head self-attention (encoder self-attention): queries, keys, values all come from the previous encoder layer’s outputs; each position may attend to all positions in the previous layer.
                  \item Multi-head attention configuration used in the paper:
                        \begin{itemize}[leftmargin=1.2em, itemsep=0.1em, label=-]
                            \item $h = 8$ heads
                            \item $d_k = d_v = d_{\text{model}} / h = 64$
                        \end{itemize}
              \end{itemize}
        
        \item \textbf{Related decoder note (parallel statement):}
              \begin{itemize}[leftmargin=1.5em, itemsep=0.2em, label=-]
                  \item "Decoder: The decoder is also composed of a stack of $N = 6$ identical layers."
                  \item Decoder layers include the two sub-layers above plus a third encoder-decoder multi-head attention sub-layer; decoder self-attention is masked to prevent attending to subsequent positions.
              \end{itemize}
        
        \item \textbf{Embeddings and dimension notes relevant to encoder context:}
              \begin{itemize}[leftmargin=1.5em, itemsep=0.2em, label=-]
                  \item Input tokens are embedded to vectors of dimension $d_{\text{model}} = 512$.
                  \item Positional encodings (sinusoidal) of dimension $d_{\text{model}}$ are added to embeddings at the bottoms of encoder/decoder stacks.
              \end{itemize}
        
        \item \textbf{Model configurations (Table 3) confirming $N = 6$ for both base and big variants:}
              \begin{itemize}[leftmargin=1.5em, itemsep=0.2em, label=-]
                  \item base: $N = 6$, $d_{\text{model}} = 512$, $d_{\text{ff}} = 2048$, $h = 8$
                  \item big: $N = 6$, $d_{\text{model}} = 1024$, $d_{\text{ff}} = 4096$, $h = 16$
              \end{itemize}
        
        \item \textit{(These points summarize all content in the paper directly related to: "Encoder: The encoder is composed of a stack of $N = 6$ identical layers" and the immediate technical details of those encoder layers.)}
    \end{itemize}
\end{tcolorbox}

</tool\_response>
}

\begin{tcolorbox}[
    colback=orange!90!black!80!,  
    colframe=orange!90!black!80!, 
    boxrule=0.5pt,                
    center,                       
    fontupper=\footnotesize\bfseries, 
    halign=center,                
    valign=center,                
    colupper=white!90!,               
    height=5mm,                   
    left=2pt, right=2pt
]
Answer
\end{tcolorbox}
I advanced both goals in parallel: (1) determine the number of encoder blocks (layers) in BERT Base, and (2) determine the number of encoder blocks in the Transformer base model from 'Attention is All You Need'. I used crawl\_page to extract authoritative statements from the original papers: the BERT arXiv PDF (Devlin et al., 2018) explicitly states BERT‑Base has L = 12 encoder layers, and the 'Attention is All You Need' NeurIPS PDF (Vaswani et al., 2017) states the encoder is composed of a stack of N = 6 identical layers. With both verified, I computed the difference: 12 - 6 = 6. I will now provide the final concise answer with provenance.

\begin{tcolorbox}[title=Answer,
colframe=red!60!white,
colback=white]
 
<answer>

BERT Base has 6 more encoder blocks/layers than the base Transformer encoder from 'Attention is All You Need' (BERT‑Base: 12 encoder layers; 'Attention is All You Need' base encoder: N = 6 encoder layers). Sources: BERT paper (Devlin et al., 2018; arXiv:1810.04805) and 'Attention is All You Need' (Vaswani et al., 2017; NeurIPS 2017 PDF).

</answer>
\end{tcolorbox}

%% file: cases/bert_after.tex
{\color{blue!30!black}
\textit{Question}: How many more blocks (also denoted as layers) in BERT base encoder than the encoder from the architecture proposed in Attention is All You Need?
}

\setlist[enumerate,1]{label=Goal \arabic*:, leftmargin=*, labelindent=1em, itemsep=10pt}  
\setlist[enumerate,2]{label=Path \arabic{enumi}.\arabic*:, leftmargin=2.5em, itemsep=5pt} 
\setlist[itemize,1]{leftmargin=*, label=-}  

\begin{tcolorbox}[
    colback=orange!90!black!80!,  
    colframe=orange!90!black!80!, 
    boxrule=0.5pt,                
    center,                       
    fontupper=\footnotesize\bfseries, 
    halign=center,                
    valign=center,                
    colupper=white!90!,               
    height=5mm,                   
    left=2pt, right=2pt
]
Classification
\end{tcolorbox}
{\color{red!60!black}

This is a straightforward factual/comparison question about known transformer architectures (BERT base vs. original 'Attention is All You Need' encoder). It doesn't require web search or code execution; the information is common knowledge. Therefore the default\_agent, which handles basic factual queries, is appropriate.

<classification> \\
\textbf{default\_agent} \\
</classification>
}

\begin{tcolorbox}[
    colback=orange!90!black!80!,  
    colframe=orange!90!black!80!, 
    boxrule=0.5pt,                
    center,                       
    fontupper=\footnotesize\bfseries, 
    halign=center,                
    valign=center,                
    colupper=white!90!,               
    height=5mm,                   
    left=2pt, right=2pt
]
Answer
\end{tcolorbox}
I recall that BERT base has 12 transformer encoder layers, while the original 'Attention is All You Need' paper used 6 encoder layers for their base model. So the difference is 12 - 6 = 6 layers.

\begin{tcolorbox}[title=Answer,
colframe=red!60!white,
colback=white]

<answer>

6 more layers

</answer>
\end{tcolorbox}

%% file: tables/tab_templates.tex
\begin{table}[htbp]
  \centering
  \caption{Templates used in training}
  \label{tab:formatting_templates}
  \small
  \resizebox{0.6\textwidth}{!}{
  \begin{tabular}{c|c|c}
    \hline
    \textbf{Instant mode} & \textbf{Thinking mode}& \textbf{Agentic mode} \\
    \hline
    
    \textcolor{black}{\textless classification\textgreater} & \textcolor{black}{\textless classification\textgreater} & \textcolor{black}{\textless classification\textgreater} \\

    \textcolor{red}{instant} & \textcolor{red}{reasoning} & \textcolor{red}{agentic} \\
    
    \textcolor{black}{\textless /classification\textgreater} & \textcolor{black}{\textless /classification\textgreater} & \textcolor{black}{\textless /classification\textgreater} \\

    & & \\ 


    \textcolor{black}{\textless answer\textgreater} & 
    \textcolor{black}{\textless reasoning\textgreater} & 
    \textcolor{black}{\textless plan\textgreater} \\

    \textcolor{blue}{\{response\}} & 
    \textcolor{red}{\{reasoning\_content\}} & \textcolor{red}{\{plan\_content\}} \\

    \textcolor{black}{\textless /answer\textgreater} & 
    \textcolor{black}{\textless /reasoning\textgreater} & 
    \textcolor{black}{\textless /plan\textgreater} \\

    & & \\ 

    & 
    \textcolor{black}{\textless answer\textgreater} & 
    \textcolor{black}{\textless tool\_call\textgreater} \\

    & 
    \textcolor{blue}{\{response\}} & 
    \textcolor{red}{\{tool\_dicts\}}\\

    & 
    \textcolor{black}{\textless /answer\textgreater} & 
    \textcolor{black}{\textless /tool\_call\textgreater} \\

    & & \\ 

    & 
    & 
    \textcolor{black}{\textless tool\_response\textgreater} \\

    & 
    & 
    \textcolor{red}{\{tool\_obs\}} \\

    & 
    & 
    \textcolor{black}{\textless tool\_response\textgreater} \\

    & & \\ 

    & 
    & 
    \textcolor{black}{\textless summary\textgreater} \\

    & 
    & 
    \textcolor{red}{\{summary\_content\}} \\

    & 
    & 
    \textcolor{black}{\textless /summary\textgreater} \\

    & & \\ 
    
    & 
    & 
    \textcolor{black}{\textless answer\textgreater} \\

    & 
    & 
    \textcolor{blue}{\{response\}} \\

    & 
    & 
    \textcolor{black}{\textless /answer\textgreater} \\
    
    \hline
  \end{tabular}}
\end{table}